%% file: root.tex
\documentclass[lettersize,journal]{IEEEtran}
\usepackage{amsmath,amsfonts}
\usepackage{algorithm}
\usepackage{algpseudocode}
\usepackage{array}
\usepackage{subfig}
\usepackage{textcomp}
\usepackage{stfloats}
\usepackage{url}
\usepackage{verbatim}
\usepackage{graphicx}
\usepackage{cite}
\usepackage{booktabs}
\usepackage{multirow}
\usepackage{diagbox}
\begin{document}






\title{DCL-SLAM: A Distributed Collaborative LiDAR SLAM Framework for a Robotic Swarm}

\author{Shipeng Zhong, Yuhua Qi$^{\star}$, Zhiqiang Chen, Jin Wu, Hongbo Chen, and Ming Liu
	\thanks{S. Zhong, Y. Qi, Z. Chen and H. Chen are with School of Systems Science and Engineering, Sun Yat-Sen University, China (Corresponding
		Author: Yuhua Qi, Email:qiyh8@mail.sysu.edu.cn)}
	\thanks{J. Wu and M. Liu are with Department of Electronic and Computer Engineering, Hong Kong University of Science and Technology, Hong Kong, China. e-mail: eelium@ust.hk}
}

\markboth{ April~2023}%
{Shell \MakeLowercase{\textit{et al.}}: A Sample Article Using IEEEtran.cls for IEEE Journals}


\maketitle

\begin{abstract}
To execute collaborative tasks in unknown environments, a robotic swarm must establish a global reference frame and locate itself in a shared understanding of the environment. However, it faces many challenges in real-world scenarios, such as the prior information about the environment being absent and poor communication among the team members. This work presents DCL-SLAM, a front-end agnostic fully distributed collaborative LiDAR SLAM framework to co-localize in an unknown environment with low information exchange. Based on peer-to-peer communication, DCL-SLAM adopts the lightweight LiDAR-Iris descriptor for place recognition and does not require full connectivity among teams. DCL-SLAM includes three main parts: a replaceable single-robot front-end LiDAR odometry; a distributed loop closure module that detects overlaps between robots; and a distributed back-end module that adapts distributed pose graph optimizer combined with rejecting spurious loop measurements. We integrate the proposed framework with diverse open-source LiDAR odometry to show its versatility. The proposed system is extensively evaluated on benchmarking datasets and field experiments over various scales and environments. Experimental result shows that DCL-SLAM achieves higher accuracy and lower bandwidth than other state-of-art multi-robot LiDAR SLAM systems. The source code and video demonstration are available at  https://github.com/PengYu-Team/DCL-SLAM.
\end{abstract}


\input{introduction}
\input{relatedwork}
\input{system}
\input{experiment}

\section{CONCLUSIONS}
\label{sec:conclusions}
This letter proposes a front-end agnostic fully distributed collaborative LiDAR SLAM framework, namely DCL-SLAM, to complete a collaborative mapping task in an unknown environment with a robotic swarm.
The proposed system is divided into three sub-modules: a flexible single-robot front-end, data-efficient distributed loop closure, and a robust back-end with minimal information exchange.
Experimental results on public and our campus datasets show the proposed framework's precision, robustness, and scalability.
In future work, we plan to explore the performance of distributed closure modules, for example, exchange data more effectively and test the system with the solid-state LiDAR more.

\bibliographystyle{IEEEtran}
\bibliography{IEEEabrv,relatepaper}

\section{Biography Section}
\vspace{-15 mm}
\IEEEaftertitletext{\vspace{-1.5\baselineskip}}
\begin{IEEEbiography}
	[{\includegraphics[width=1in,height=1.25in,clip,keepaspectratio]{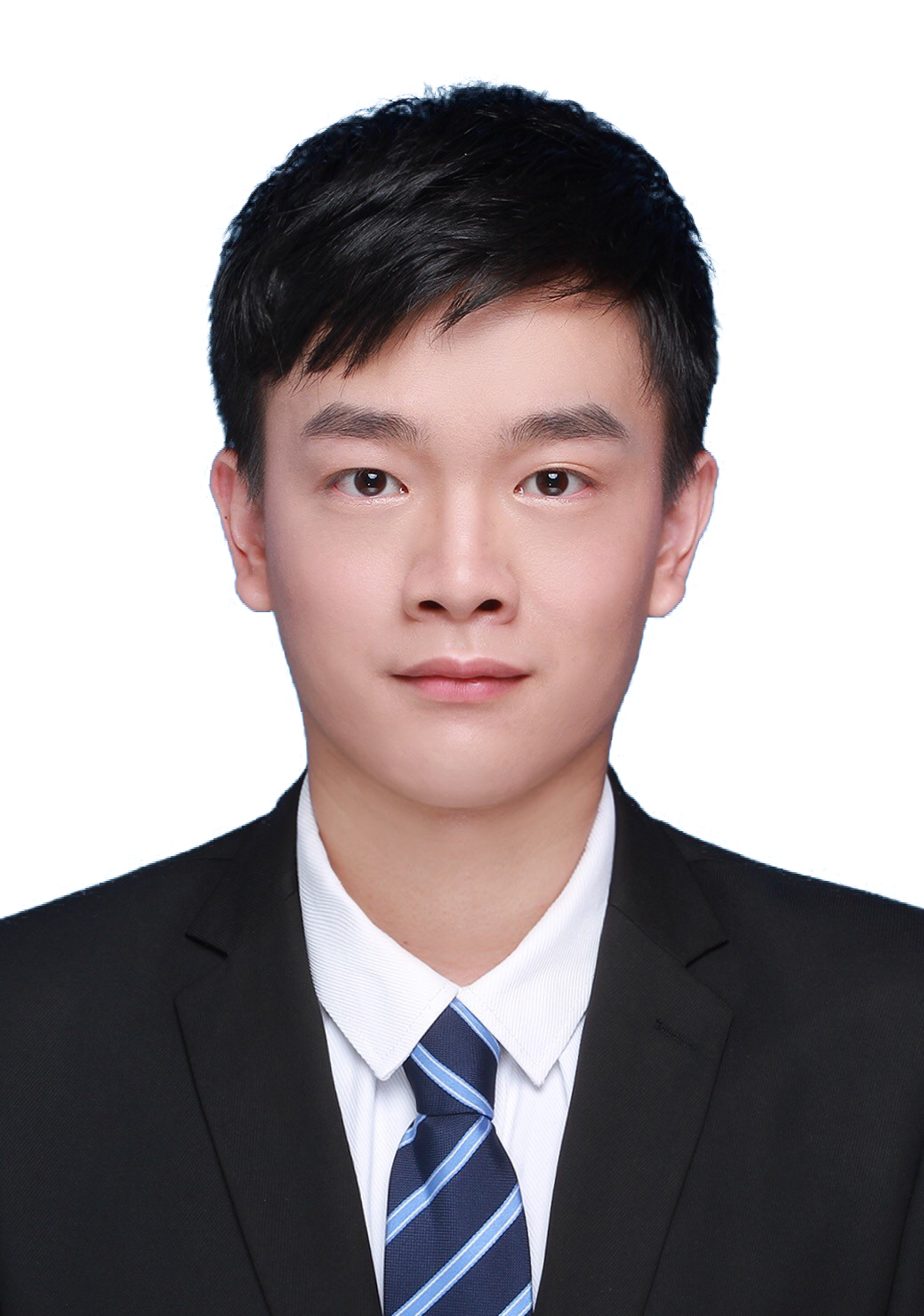}}]{Shipeng Zhong} received his B.E. degree in the School of Computer Science and Engineering from Sun Yat-sen University, Guangzhou, Guangdong, China, 2020. He is currently a Ph.D. student in the School of Systems Science and Engineering at SYSU under the instruction of Prof. Hongbo Chen. His research interests are in the areas of autonomous unmanned systems and cooperative mapping.
\end{IEEEbiography}
\vspace{-12 mm}
\begin{IEEEbiography}
	[{\includegraphics[width=1in,height=1.25in,clip,keepaspectratio]{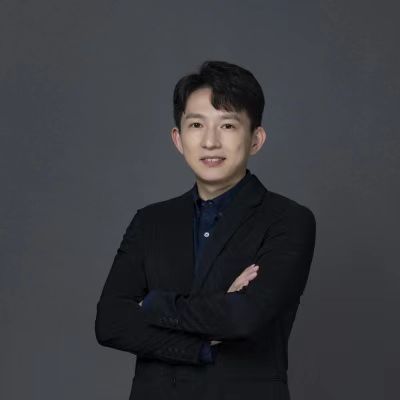}}]{Yuhua Qi} is currently a Postdoctoral Researcher in the School of Systems Science and Engineering at Sun Yat-Sen University, Guangzhou, China. He received his B.S. and Ph.D. in Aerospace Engineering at Beijing Institute of Technology, Beijing, China. His research interests include multi-robot SLAM, cooperative control and autonomous unmanned systems.
\end{IEEEbiography}
\vspace{-12 mm}
\begin{IEEEbiography}[{\includegraphics[width=1in,height=1.25in,clip,keepaspectratio]{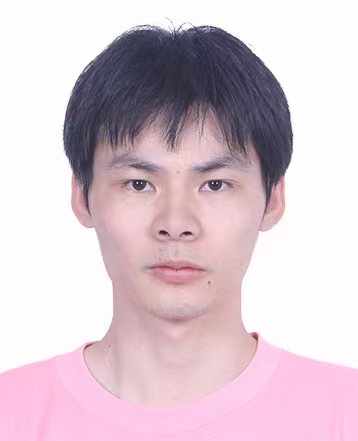}}]{Zhiqiang Chen} is currently a graduate student in the School of Systems Science and Engineering at Sun Yat-Sen University, Guangzhou, China. He received his B.E. in Sun Yat-sen University, Guangzhou, China. His research interests include LiDAR SLAM and sensor fusion.
\end{IEEEbiography}
\vspace{-12 mm}
\begin{IEEEbiography}[{\includegraphics[width=1in,height=1.25in,clip,keepaspectratio]{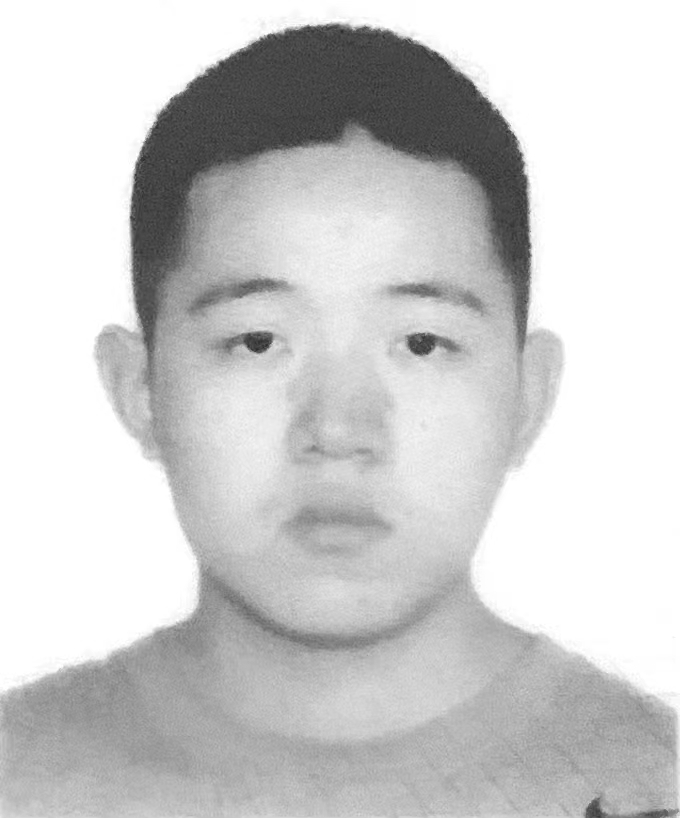}}]{Jin Wu} (Member, IEEE) was born in May, 1994, in Zhenjiang, Jiangsu, China. He received the B.S. degree from the University of Electronic Science and Technology of China, Chengdu, China. He is currently a PhD student with RAM-LAB, HKUST, under the supervision of Prof. Ming Liu. In 2013, He was a visiting student in Groep T, Katholieke Universiteit Leuven. From 2019 to 2020, He was with Tencent Robotics X. He is currently study pose estimation in robotics.
\end{IEEEbiography}
\vspace{-12 mm}
\begin{IEEEbiography}[{\includegraphics[width=1in,height=1.25in,clip,keepaspectratio]{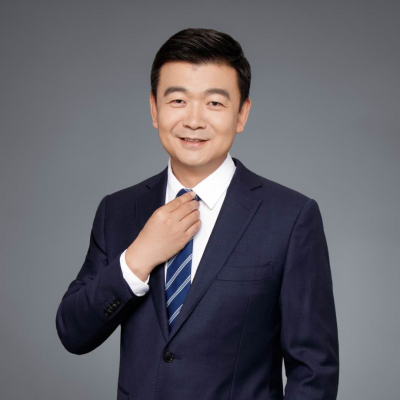}}]{Chen Hongbo} graduated from the School of Astronautics of Harbin Institute of Technology, and obtained a Ph.D in aircraft design and engineering. He started his career in the First Research Institute of China Aerospace Science and Technology Corporation (CASC) in 2007; in 2018, he was transferred to the School of Systems Science and Engineering, Sun Yat-Sen University. Chen Hongbo has abundant work experience in the core positions of the aerospace industry. He has been engaged in the pre-research and innovation work of major national projects for years. His innovative research and system engineering practices concentrate in the fields of aerospace transportation systems and aerospace unmanned systems, many key technical problems have been conquered by him and his team.As the main author, Chen Hongbo has been granted 26 patents, including 11 inventions; as the first author or main author, he has published more than 30 academic papers and 2 books.
\end{IEEEbiography}
\vspace{-8 mm}
\begin{IEEEbiography}[{\includegraphics[width=1in,height=1.25in,clip,keepaspectratio]{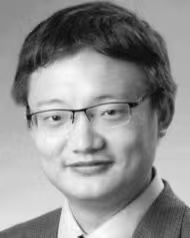}}]{Ming Liu} (Senior Member, IEEE) received the B.A. degree in automation from Tongji University, Shanghai, China,
	in 2005, and the Ph.D. degree from the Department of Mechanical and Process Engineering, ETH Zurich, Zurich, Switzerland, in 2013, supervised by
	Prof. Roland Siegwart. During his master's study with Tongji University, he stayed one year with the Erlangen-Nunberg University and Fraunhofer Institute IISB, Erlangen, Germany, as a Master Visiting Scholar.\\
	\indent He is currently with the Electronic and Computer Engineering, Computer Science and Engineering Department, Robotics Institute, The Hong Kong University of Science and Technology (HKUST), Hong Kong, China, as an Associate Professor. He is also a founding member of Shanghai Swing Automation Ltd., Co. He is currently the chairman of Shenzhen Unity Drive Inc., China. He is coordinating and involved in NSF Projects and National 863-Hi-TechPlan Projects in China. His research interests include dynamic environment modelling, deep-learning for robotics, 3-D mapping, machine learning, and
	visual control.\\
	\indent Dr. Liu was a recipient of the European Micro Aerial Vehicle Competition
	(EMAV'09) (second place) and two awards from International Aerial Robot
	Competition (IARC'14) as a Team Member, the Best Student Paper Award
	as first author for MFI 2012 (IEEE International Conference on Multisensor
	Fusion and Information Integration), the Best Paper Award in Information
	for IEEE International Conference on Information and Automation (ICIA
	2013) as first author, the Best Paper Award Finalists as co-author, the Best
	RoboCup Paper Award for IROS 2013 (IEEE/RSJ International Conference
	on Intelligent Robots and Systems), the Best Conference Paper Award for
	IEEE CYBER 2015, the Best Student Paper Finalist for RCAR 2015 (IEEE
	International conference on Real-time Computing and Robotics), the Best
	Student Paper Finalist for IEEE ROBIO 2015, the Best Student Paper for IEEE ROBIO 2019, the Best Student Paper Award for IEEE ICAR 2017, the Best Paper in Automation Award for IEEE ICIA 2017, twice the innoviation contest Chunhui Cup Winning Award in 2012 and 2013, and the Wu Wenjun AI Award in 2016. He was the Program Chair of IEEE RCAR 2016 and the Program Chair of International Robotics Conference in Foshan 2017. He was the Conference Chair of ICVS 2017. He has published many popular papers in top robotics journals including \textsc{IEEE Transactions on Robotics}, \emph{International Journal of Robotics Research} and \textsc{IEEE Robotics and Automation Magazine}. He is a program member of Robotics: Science and Systems (RSS) 2021.\\
	\indent Dr. Liu is currently an Associate Editor for \textsc{IEEE Robotics and Automation Letters}, \emph{International Journal of Robotics and Automation}, \emph{IET Cyber-Systems and Robotics}, IEEE IROS Conference 2018, 2019 and 2020. He served as a Guest Editor of special issues in \textsc{IEEE Transactions on Automation Science and Engineering}. He is a Senior Member of IEEE.
\end{IEEEbiography}

\vfill

\end{document}

%% file: introduction.tex
\section{INTRODUCTION}
\IEEEPARstart{S}{imultaneous} Localization and Mapping (SLAM) is a fundamental capability in robot navigation, especially in unknown and GPS-denied environments, and there is a substantial body of literature dedicated to advanced single-robot SLAM methods \cite{slamReview_cadena2016,shan2020_liosam,he2020integrated,xu2021_fastlio2,guo2022lidar,wang2022feature}. However, compared to a single-robot system, the multi-robot system has a more remarkable ability and work efficiency in time-sensitive applications such as factory automation, exploration of unsafe areas, intelligent transportation, and search-and-rescue in military and civilian endeavors. Hence, various collaborative SLAM methods (C-SLAM) for robotic swarms have been studied in the last decade based on single-robot SLAM \cite{lajoie2021_cslamreview}. C-SLAM aims to combine data from each robot and establish relative pose transformations and a global map between robots.


\begin{figure}[thpb]
	\centering
	\includegraphics[width=0.99\linewidth]{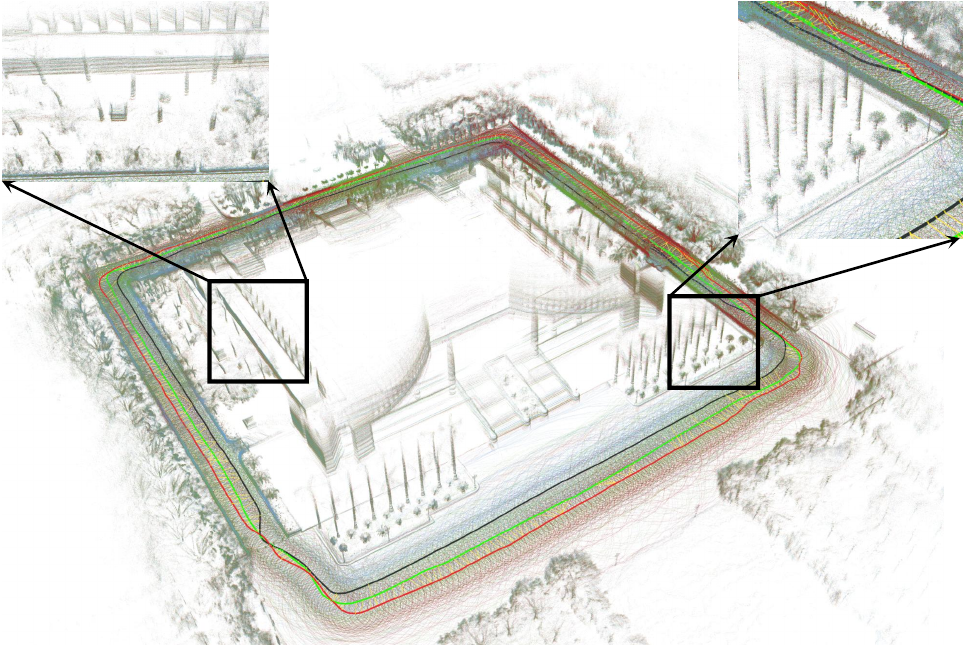}
	\caption{The reconstructed maps and trajectories derived by DCL-SLAM framework with three robots, shown in green, red and blue, at Sun Yat-Sen University's eastern campus library.}
	\label{fig:library}
\end{figure}

For the real-world application of robotic swarms, however, C-SLAM remains an open problem and faces several challenges currently being researched since communication among the robots may be slow (bandwidth constraints) or infeasible (limited communication range). Due to the various advantages of LiDAR, such as high-resolution point cloud and robustness to a wide range of weather and lighting conditions, many LiDAR C-SLAM methods have been proposed \cite{ebadi2020_lamp, Chang2022LAMP2A, dube2020_segmap, zhu2021paircon, huang2021_discoslam}.
However, these systems do not apply to a robotic swarm, especially in a large-scale scenario with communication constraints.
In addition, many C-SLAM works \cite{dube2020_segmap,huang2021_discoslam} evaluate their system by splitting a dataset into multiple parts, which cannot represent the actual cooperation of robotic swarms. Therefore, gathering synthetic data from multiple platforms\cite{leung2011utias} in real scenarios for system evaluation is necessary. C-SLAM is still a relatively young research field in the research community, albeit very promising.



In this spirit, we present DCL-SLAM, a fully distributed collaborative LiDAR SLAM framework for robotic swarms. DCL-SLAM focuses on data-efficient and resilient inter-robot communication, reaching a consistent trajectory estimate with partial pose measurements and indirect data association. The framework is compatible with various LiDAR odometry based on the requirements and the scenario with no additional effort. Each robot performs single-robot front-end independently and co-localizes when teammates are within communication range. To satisfy the communication constraints, a lightweight global descriptor, LiDAR-Iris \cite{wang20K20_iris}, is integrated to describe laser scans and detect loop closures without exchanging extra raw data.
What's more, we design a three-stage data-efficient distributed loop closure approach to obtain relative pose transformations between robots.
After that, outlier rejection based on \cite{Mangelson2018_pcm} is performed to remove spurious inter-robot loop closures for robustness.
Finally, all collected measurements are optimized together by a two-stage distributed Gauss-Seidel (DGS) approach \cite{Choudhary16icra}, which requires minimal information exchange.
Our framework is evaluated using several multi-robot datasets, including public datasets (KITTI) and campus datasets gathered with three unmanned ground vehicles (UGVs). Furthermore, online field test and scalability experiment are also presented.

To summarize, the contributions of this work are the following:
\begin{itemize}
	\item A front-end agnostic fully distributed collaborative LiDAR SLAM framework for the robotic swarm is presented, which supports rapid migration to different LiDAR sensors, platforms, and scenarios.
	\item A highly-integrated data-efficient inter-robot loop closure detection approach is proposed to achieve high accuracy and robustness with a three-stage communication pipeline.
	\item Extensive experimental evaluation, including performance, scalability and real-world field tests, is conducted to validate the proposed system. The custom datasets and the source code of the proposed system are open to the public. 
\end{itemize}

The rest of this paper is organized as follows.
After reviewing the relevant inter-robot loop closure, outlier rejection, multi-robot pose graph optimization background literature in Section \ref{sec:related_work}, a framework for the proposed system is presented in Section \ref{sec:system_architecture}.
Experimental results are given in Section \ref{sec:experiment}, with conclusions in Section \ref{sec:conclusions}.

%% file: relatedwork.tex
\section{RELATED WORK}

\label{sec:related_work}
$\mathbf{Multi\text{-}robot\ Loop\ Closure.}$
Estimating the relative pose between the robots by detecting inter-robot loop closures is an efficient way to merge the trajectories of the robots in a common frame without external positioning infrastructure. What's more, it is also critical to compensate for the drift of the front-end odometry and improve the accuracy of trajectory estimates. The first issue is how to detect inter-robot loop closures effectively. In  \cite{ebadi2020_lamp, Chang2022LAMP2A, dube2017_mrslam}, the local laser maps produced by each robot were directly transmitted to a server for centralized map fusion, which requires high computation resources on the server and reliable communication performance. 
Many features-based visual C-SLAM \cite{cvislam_karrer2018, schmuck2019_ccmslam, lajoie2020_doorslam, kimera_chang2021, zhu2021paircon, ouyang2021collaborative, schmuck2021_covins}
shared the binary descriptor (e.g., ORB and BRIEF) of the keyframes and used a bag-of-words (BoWs) approach to find multi-robot loop closures. 
PairCon-SLAM \cite{zhu2021paircon} proposed distributed, online, and real-time SLAM algorithm to construct dense maps of large-scale scenarios with LiDAR or RGBD camera in two personal computers (PCs), but it was not suitable for robotic swarms.
In addition, \cite{shan2021robust} used projected LiDAR intensity image to extract ORB features for place recognition using the BoWs method. Introducing the keyframe global descriptor with BoWs enables fast and robust searching for loop closures. Moreover, it reduces the system's bandwidth and the computation at the back-end optimization, which has a certain reference significance for LiDAR C-SLAM. Generally, descriptor extraction on the point cloud can be divided into local, global, and learning-based descriptors. Local descriptor \cite{3dsurf_knopp2010, bshot_prakhya2015} was extracted from the local neighborhood of the keypoint in the point cloud. Although local descriptors methods are unsuitable for inter-robot loop closure detection scenarios due to the lack of descriptive power, they could be used to validate putative loop closures and estimate relative pose in geometric verification.
Extracting global descriptors \cite{delight_cop2018, wang20K20_iris, he2016_m2dp, kim2018_scancontext} from the entire set of points can mitigate the problem of lack of descriptive power.
DiSCo-SLAM \cite{huang2021_discoslam} introduced Scan-Context \cite{kim2018_scancontext} for inter-robot loop closure and proposed a two-stage local and global distributed optimization framework.
Compared to DiSCo-SLAM, considering bandwidth limitation and communication range, our method propose a distributed loop closure framework with a three-stage communication pipeline, which avoids exchanging all raw or feature clouds. Second, we conduct extensive and comprehensive experiments to validate the proposed system and apply our system in a large-scale team with nine robots while only three in DiSCo-SLAM.
Recently, Convolutional neural networks have been used to train feature descriptors \cite{qi2017pointnet, yew20183dfeat,fischer2021stickypillars,li2021ssc,ye2022efficient}.
SegMap \cite{dube2020_segmap} proposed a data-driven descriptor for LiDAR to extract meaningful features that can be used for loop detection and map reconstruction. 
FPET-Net\cite{ye2022efficient} introduced a feature point extraction module to reduce the size of the point cloud and preserve the data features, and a point transformer module to extract the global descriptors.
However, these learning-based approaches require tremendous amounts of training data. In our work, we adopted a lightweight global descriptor LiDAR-Iris for fast and robust distributed loop closure detection. It encodes the height information of each bin, extracts discriminative features without any pre-training, and is rotation-invariant without brute-force matching.

Obtaining the relative pose of loop closure is the second issue. The 3D-3D matching approach is usually adopted in LiDAR SLAM, such as ICP \cite{icp_besl1992} and GICP \cite{koide2021voxelized}. 
How to exchange the descriptors on each robot is a critical problem due to the communication range and bandwidth limitation in the distributed system.
\cite{cieslewski2017efficient} pre-assigned words of the vocabulary to each robot and sent the entire query to the candidate robot to detect loop closures.
\cite{cieslewski2017_ddvslam} performed efficient data association for distributed loop closure with NetVLAD descriptor in a fully connected team.
These approaches require a smaller amount of data exchange and scale well with the number of robots in the team, but are designed for the full-connected system.
\cite{choudhary2017_dgs, wang2019active, lajoie2020_doorslam} efficiently exchange data during robots' rendezvous, accounting for the available communication and computation resources.
These approaches are more suitable for robotic swarms, especially in large-scale scenarios with limited communication range.

\begin{figure*}[h]
	\centering
	\includegraphics[width=0.88\linewidth]{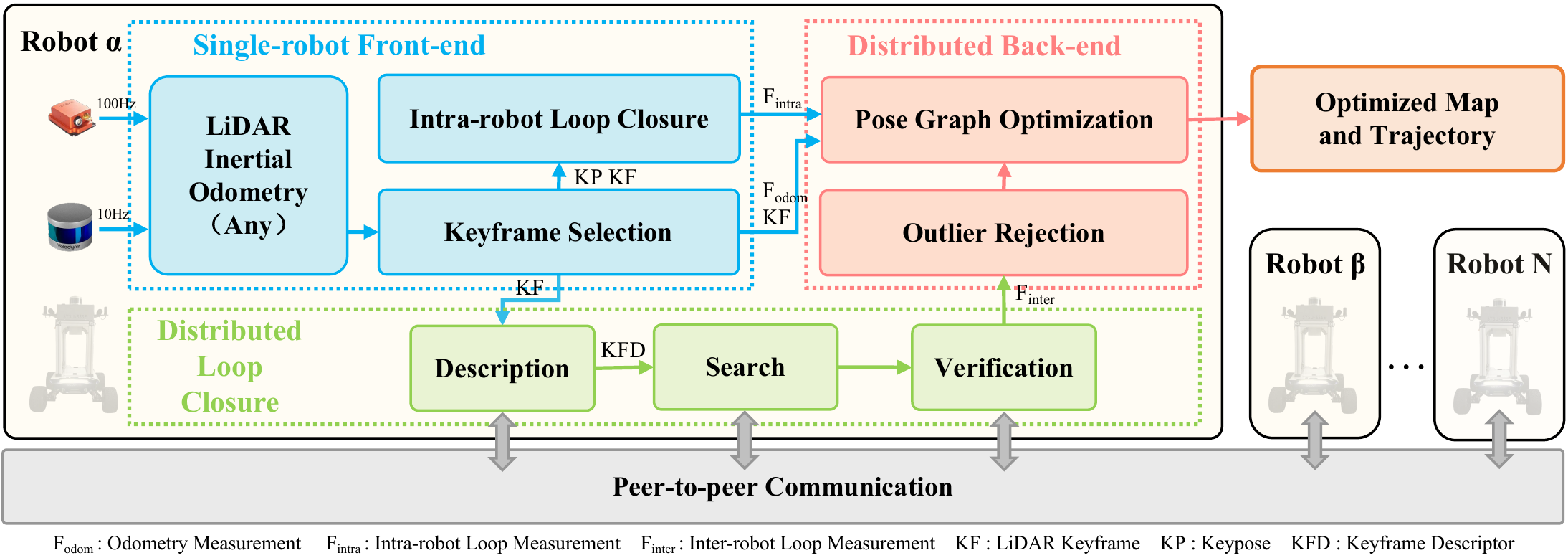}
	\caption{The proposed fully-distributed collaborative LiDAR SLAM, DCL-SLAM, consists of a replaceable single-robot front-end module to produce odometry factor and intra-robot loop factor, a distributed loop closure module for detecting inter-robot loop closures, a distributed pose graph optimization module to estimate the global trajectories for the robotic swarm combined with rejecting spurious loop measurements, and a peer-to-peer communication module to exchange necessary data (descriptors, measurements, pose estimates, etc.).}
	\label{fig:systemoverview}
\end{figure*}

$\mathbf{Outlier\ Rejection.}$
Outlier rejection is critical to pose estimation since incorrect measurements may be caused by perceptual aliasing.
The RANSAC algorithm \cite{fischler1981random}, which is widely used in both LiDAR and visual SLAM, is robust to a large portion of outliers.
However, it distinguishes inliers and outliers based on a given model.
\cite{do2020robust} rejects inaccurate loop closures by considering the data similarity in the measurement and the consistency between loop closures.
\cite{Mangelson2018_pcm} proposes pairwise consistent measurement set maximization (PCM) to check the consistency of each pair of inter-robot loop closures to merge multi-robot map robustly, which several multi-robot SLAM systems have adopted \cite{huang2021_discoslam, lajoie2020_doorslam, kimera_chang2021, ouyang2021collaborative} for outlier rejection.
This work combines RANSAC and PCM to make the system more robust.



$\mathbf{Multi\text{-}robot\ Pose\ Graph\ Optimization.}$
Pose Graph Optimization is one of the popular and effective methods to optimize poses in the back end.
Centralized PGO approaches \cite{dube2017_mrslam, rosen2019se} collect all information at a server or robot and estimate the trajectory for all the robots on a global pose graph.
Still, distributed approach \cite{ddfsam_alexander2013, Choudhary16icra} is more suitable for a multi-robot system under poor communication due to the bandwidth constraints and limited communication range.
DDF-SAM \cite{ddfsam_alexander2013} proposed a landmark-based distributed trajectory optimization method based on Gaussian elimination.
\cite{tron2016distributed} propose a multi-stage distributed Riemannian consensus protocol for distributed execution of Riemannian gradient descent.
\cite{Choudhary16icra} proposed a two-stage distributed Gauss-Seidel method and avoided the complex bookkeeping and double counting.
\cite{fan2020majorization} proposed a majorization-minimization approach to solve distributed PGO.
\cite{tian2021distributed} proposes Riemannian Block-Coordinate Descent (RBCD) to solve a rank-restricted relaxation of the pose graph optimization problem and reach a better performance than DGS with a local network.
Distributed C-SLAM solutions \cite{lajoie2020_doorslam, cieslewski2017_ddvslam} have adopted \cite{Choudhary16icra} since their applicability to limited communication resources.
In this work, we also use the DGS method as a back-end for the experiments.

%% file: system.tex
\section{THE DCL-SLAM FRAMEWORK}
\label{sec:system_architecture}
Our proposed framework relies on peer-to-peer communication with the ad-hoc wireless network. When two robots rendezvous, they share the understanding of the environment information to establish a global reference frame and perform distributed SLAM. Moreover, each robot performs single-robot SLAM when there is no teammate within the communication range to ensure the independence of each robot.

In some existing C-SLAM systems \cite{dube2020_segmap}, the front-end and back-end modules are tightly entangled, making it difficult to replace or update one improved module (e.g., odometry, inter-robot loop closure) at a low-cost. Therefore, our framework is modularized into three main parts: a single-robot front-end, distributed loop closure, and distributed back-end. An overview of our proposed DCL-SLAM framework is given in Fig. \ref{fig:systemoverview}. Each robot collects raw data from a LiDAR and an Inertial Measurement Unit (IMU) and locally runs a \textit{single-robot front-end} module (Section \ref{sec:local_front_end}) to produce an estimate of its trajectory.
In \textit{distributed loop closure} module (Section \ref{sec:distributed_loop_closure}), a keyframe scan is selected from odometry pose and then described as a global descriptor (LiDAR-Iris \cite{wang20K20_iris}) for performing place recognition among the robots. Each robot communicates to other robots within the communication range with the descriptor for performing inter-robot loop closure detection. The output loop closure candidates are then verified using either the raw point cloud or the features point cloud with ICP method and random sample consensus (RANSAC) algorithm to get inter-robot loop closure measurements.  
After that, the \textit{distributed back-end module} module (Section \ref{sec:distributed_pose_graph_optimization}) module collects all odometry measurements, intra-robot loop measurements, and inter-robot loop measurements to compute the maximal set of pairwise consistent measurements (PCM) \cite{Mangelson2018_pcm} and filters out the outliers.
The optimal trajectory estimation of the robotic swarm is solved using a two-stage distributed Gauss-Seidel method proposed in \cite{Choudhary16icra} for reaching an agreement on the pose graph configuration.


\subsection{Single-robot Front-end}
\label{sec:local_front_end}
The DCL-SLAM framework support interface with single-robot front-ends with various LiDAR odometry and is compatible with different LiDAR configurations (e.g., solid-state and mechanical LiDAR), platforms (e.g., ground and aerial vehicles), and scenarios (e.g., indoor and outdoor environment).
In our implementation, LIO-SAM \cite{shan2020_liosam} is adopted as a typical example of odometry for ground vehicles, FAST-LIO2 \cite{xu2021_fastlio2} for aerial vehicles, and other LiDAR odometry such as LOAM \cite{zhang2014_loam} is an alternative.
Once the odometry estimate is refined, the LiDAR odometry factor $F_{odom}^{r}$ ($r \in\{\alpha,  \beta,  \gamma, \cdots\}$) can be constructed and sent to the distributed back-end module. The correlative point cloud scan is used for intra-robot and inter-robot loop closure.

To improve the accuracy of the single-robot front-end module, an intra-robot loop closure submodule to eliminate the odometry drift is necessary. In some existing LiDAR odometry (e.g., LIO-SAM), the intra-loop closure submodule is included. Therefore, it is easy to extract the results to construct the intra-robot loop closure factor $F_{intra}^{r}$ for the pose distributed graph optimization. In addition, we also provide a built-in intra-robot loop closure submodule in our framework for only odometry systems (e.g., FAST-LIO2), familiar with the inter-robot loop closure module without communication with other robots (see details in the following subsection).

\subsection{Distributed Loop Closure}
\label{sec:distributed_loop_closure}
\begin{figure}[ht]
	\centering
	\includegraphics[width=0.99\linewidth]{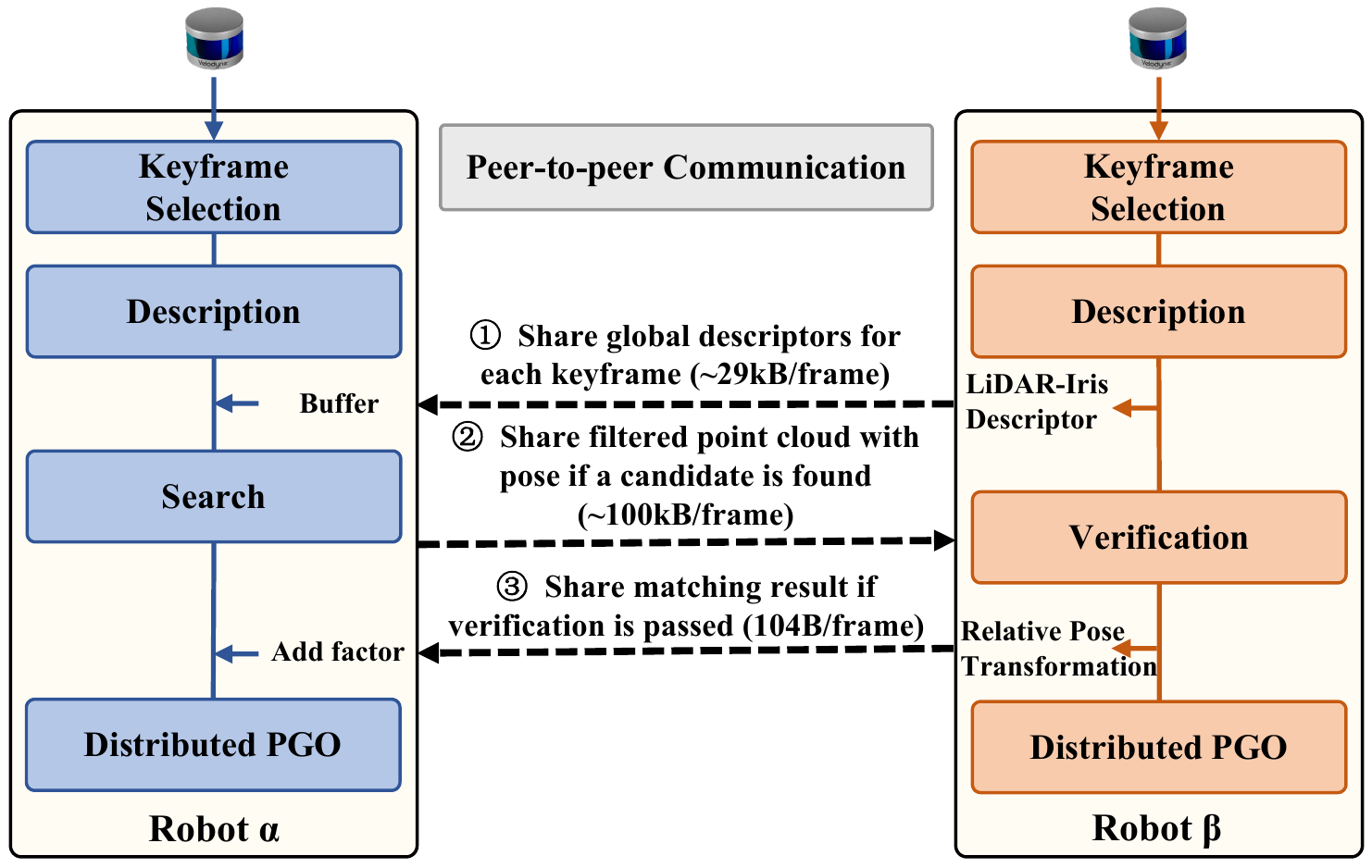}
	\caption{Illustration of the distributed loop closure module from robot $\alpha$'s perspective (using VLP-16).}
	\label{fig:distributedLoopFramework}
\end{figure}

In our framework, the distributed loop closure module performs place recognition among robotic swarms to generate the relative poses transformation and is illustrated in Fig. \ref{fig:distributedLoopFramework}. Moreover, detecting overlapping parts between local maps created by itself or other robots in the swarm is crucial to reduce its accumulated drift and fuse all robots into a geometrically consistent global map.
Our distributed loop closure module consists of four steps: keyframe selection, description, search, and verification. 
To improve the computation and communication efficiency, we first select keyframes when the position and rotation change compared to the previous exceeds a threshold. In our work, the position and rotation change thresholds are 1 meter and 0.2 rad. Furthermore, we perform a three-stage distributed loop closure detection approach, including sharing global descriptors for each keyframe, sharing filtered point cloud with pose if a candidate is found, and sharing matching result if verification pass. 
It can permit data-efficient communication and does not require full connectivity among the robotic swarm. 

\begin{figure}[h]
	\centering
	\includegraphics[width=0.62\linewidth]{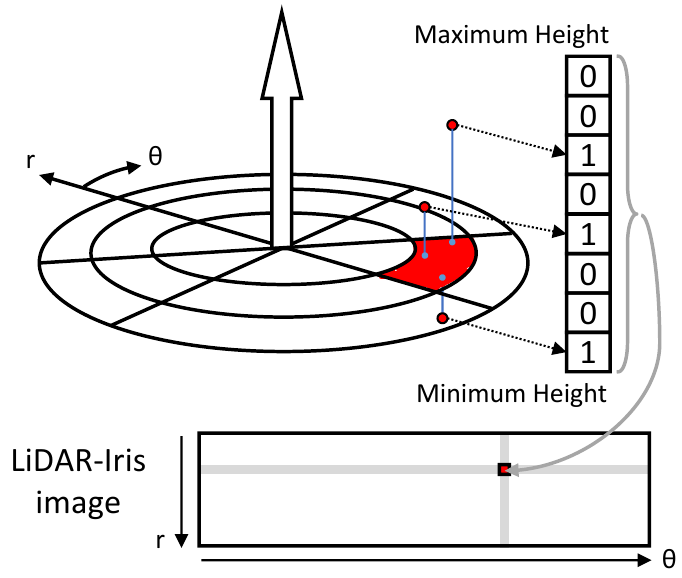}
	\caption{Demonstration of encoding the laser scan into the global descriptor.}
	\label{fig:iris}
\end{figure}

$\mathbf{Description.}$
LiDAR-Iris\cite{wang20K20_iris}, a state-of-the-art global LiDAR descriptor, is adopted for multi-robot place recognition. 
As shown in Fig. \ref{fig:iris}, laser scan $S$ is projected to its bird-eye view and divides into $N_r$ (radial direction) $\times$ $N_a$ (angular direction) bins according to the angular and radial resolution, defined by $S = \bigcup S_{ij}, i \in [1,2,...,N_r], j \in [1,2,...,N_a]$.
Each bin $S_{ij}$ encodes height information into an 8-bit binary code $c_{ij}$ as the pixel intensity of the image $I$:
\begin{equation}
I = (c_{ij}) \in \mathbb{R}^{N_r\times N_a}.
\end{equation}
Four 1D LoG-Gabor filters are exploited to convolve each row of image $I$ to extract the image features. The 1D LoG-Gabor filter has the following form in the frequency domain:
\begin{equation}
G(\omega) = \exp(-\frac{\ln^2(\omega/\omega_0)}{2\ln^2(\sigma/\omega_0)}),
\end{equation}
where $\omega_0$ is filter's centre frequency the and $\sigma$ is the parameter of the filter bandwidth.
A discriminative binary feature image is obtained by applying a simple thresholding operation on convolutional responses.
This description procedure does not require counting the points in each bin and pre-train model.

In this paper, for Velodyne HDL-64E (KITTI dataset) and VLP-16 (our dataset), $N_r$ is 80, and $N_a$ is 360.
In each robot, the scan will be locally described as a binary feature image.
As shown in Fig. \ref{fig:distributedLoopFramework}, once robot $\beta$ is in the communication range of robot $\alpha$, the binary feature images of robot $\beta$ are sent to robot $\alpha$ alone with row keys (about 29kB per keyframe).
Therefore, robot $\alpha$ can obtain a history database of binary feature images for all robots.

$\mathbf{Search.}$
The recipient robot (robot $\alpha$) searches for neighbors of the shared binary feature images.
To deal with the viewpoint changes (e.g., revisit in the opposite direction), a Fourier transformation is adopted to estimate the rotation between two images. Suppose that the images $I_\alpha$ and $ I_\beta$ have a shift $(\Delta m,\Delta n)$, then its Fourier transformation $F_\alpha(u,v)$ has the following properties: 
\begin{equation}
\begin{aligned}
F_\alpha(u,v) &= F\{I_\alpha(m,n))\}\\ &= F\{I_\beta(m-\Delta m,n-\Delta n))\}\\ &= \exp(-i(u\Delta x + v\Delta y) F_\beta(u,v),
\end{aligned}
\end{equation}
which meaning:
\begin{equation}
\begin{aligned}
(\Delta m,\Delta n) &= \underset{m,n}{\arg\max}\ \exp(-i(u\Delta m + v\Delta n)\\ &= \underset{m,n}{\arg\max}\frac{F_\beta(u,v)}{F_\alpha(u,v)}.
\end{aligned}
\end{equation}
Since the distance calculation on all binary feature images in the history database is heavy, we use a hierarchical nearest search by introducing a row key to achieve a fast search. Each robot further reduces the image to a $N_r$ $\times$ 1 vector as query row key $K$ by encoding each image row into a single value via a $L_2$ norm:
\begin{equation}
\begin{aligned}
K &= (k_i) \in \mathbb{R}^{N_r},\\ k_i &= (\sum_{j=1}^{N_a}\mid c_{ij}\mid^2)^{\frac{1}{2}},
\end{aligned}
\end{equation}
where $c_{ij} \in I$. The row key $K$ is rotation-invariant because the encoding function is independent of the viewpoint.
In the search stage, a row key kd-tree is built for loop closure candidate search, and the neighbors' binary feature images are compared against the query binary feature image using Hamming distance:
\begin{equation}
d = \sum^{N_r}\sum^{N_a} b^{\alpha}_{ij}\oplus b^{\beta}_{ij} < \eta,
\end{equation}
where $b^{\alpha}_{ij}$ is the binary feature of $c^{\alpha}_{ij} \in I^{\alpha}$, $\eta$ is the distance threshold and $\oplus$ is XOR operation.
The closest candidate to the query satisfying an acceptance threshold is regarded as a loop-closure candidate and shared with robot $\beta$.
Robot $\alpha$ also transmitted the features cloud or deskewed cloud of the recognized place to robot $\beta$ for verification (about 100kB per frame).
This stage provides a loop closure candidate without exchanging the raw data and does not require full connectivity.

$\mathbf{Verification.}$
In the verification stage, the candidate matching items are then verified by a scan-to-map matching performed on the filtered scan of robot $\alpha$ and the surrounding submap of robot $\beta$ using the RANSAC algorithm.
If the set of inliers is sufficiently large, it considers the corresponding loop closure successful.
The relative pose transformation of the loop closure candidate is evaluated using the ICP method.
After that, robot $\beta$ feeds back the relative pose transformation of the inter-robot loop closure to robot $\beta$ (about 104B per keyframe).
Once the inter-robot loop factor $F_{inter}^r$ is found and shared, both robots initiate the distributed pose graph optimization described in the following section.

\begin{algorithm}[t]
	\caption{Collaborative localization}\label{alg:alg1}
	\renewcommand{\algorithmicrequire}{\textbf{Input:}}
	\renewcommand{\algorithmicensure}{\textbf{Output:}}
	\begin{algorithmic}[1]
		\Require
		\Statex LIO odometry: $F_{odom}$
		\Statex Intra robot loop closure: $F_{intra}$
		\Statex Inter robot loop closure: $F_{inter}$
		\Ensure 
		\Statex Full poses of trajectories: $x$
		
		\While {!\emph{graph\_is\_connected()}}
		\State \emph{sleep()};
		\EndWhile
		
		\State connected\_robot $\gets$ \emph{compute\_optimization\_order($F_{inter}$)};
		\State neighbor\_robot $\gets$ \emph{check\_robot\_state(connected\_robot)};
		\For {$i=1$ to optimization\_iteration\_time}
		\For {$r\ \in\ $neighbor\_robot}
		\State $G^r\ \gets$ \emph{create\_pose\_graph($F_{odom}^r$,$F_{inter}^r$,$F_{intra}^r$)};
		\State \emph{send\_start\_flag\_and\_initialization()};
		\If{!all\_robot\_start}
		\State \emph{stop\_optimization()};
		\EndIf
		\State $F_{inter}^r$ $\gets$ \emph{outlier\_rejection($F_{odom}$,$F_{inter}^r$)};
		\State $R^r \gets$ \emph{estimate\_rotation($G^r, R^{other\_robot}$)};
		\State \emph{shared\_correlative\_rotation\_estimate($R^r$)};
		\If{!all\_robot\_finish\_rotation\_estimation}
		\State \emph{stop\_optimization()};
		\EndIf
		\State $x^r\ \gets$ \emph{estimate\_pose($G^r,x^{other\_robot}$)};
		\State \emph{shared\_correlative\_pose\_estimate($x^r$)};
		\If{!all\_robot\_finish\_pose\_estimation}
		\State \emph{stop\_optimization()};
		\EndIf
		\EndFor
		\EndFor
		\State \Return $x = \bigcup x^r, r\in\{\alpha, \beta,\cdots\}$
	\end{algorithmic}
	\label{alg1}
\end{algorithm}
\subsection{Distributed Back-end}
\label{sec:distributed_pose_graph_optimization}
In our DCL-SLAM framework, the distributed back-end module includes two submodules. The first submodule, outlier rejection, removes spurious inter-robot loop closures. The second submodule, pose graph optimization (PGO), constructs a global pose graph with all factors to estimate the global trajectories for the robotic swarm.
Three types of factors are considered in back-end PGO, including odometry factor $F_{odom}^r$, intra-robot loop factor $F_{intra}^r$, and inter-robot loop factor $F_{inter}^r$. A general likelihood item for these factors is:
\begin{equation}
\phi(x) = \psi (z_{\alpha_i\beta_j}\mid x),
\end{equation}
where $x = [x_{\alpha}, x_{\beta}, x_{\gamma}, \cdots]$ is a set of poses of trajectories of robots and $z_{\alpha_i\beta_j}$ is a general measurement model related to the pose transformation of robot ${\alpha}$ at time $i$ and that of robot ${\beta}$ at time $j$.


$\mathbf{Outlier \ Rejection.}$
The distributed outlier rejection submodule rejects false-positive inter-robot loop closures caused by only matching features in the loop finding procedure. 
These perceptual aliasing may cause a considerable drift in the robot trajectory estimation.
Therefore, pairwise consistent measurements set maximization in \cite{Mangelson2018_pcm} is introduced for outlier rejection and implemented as a distributed approach.
PCM checks the consistency of pairs of inter-robot loop closure measurements $z_{\alpha_j \beta_m}$ and $z_{\alpha_i \beta_n}$ by the following metric:
\begin{equation}
|| (z_{\alpha_i \alpha_j} \cdot z_{\alpha_j \beta_m} \cdot z_{\beta_m \beta_n}) \cdot z_{\alpha_i \beta_n}^{-1} ||_2^2 < \epsilon,
\end{equation}
where $z_{\alpha_i \alpha_j}$ is a intra-robot pose transformation related to time $i$ and $j$ in robot $\alpha_i$, $z_{\alpha_j \beta_m}$ is a inter-robot pose transformation related to time $j$ in robot $\alpha$ and $m$ in robot $\beta$, and $\epsilon$ is the likelihood threshold.
After the pairwise consistency checks are performed, it finds the largest set of pairwise consistent measurements by finding a maximum clique.
Finally, the loop closures in the largest set are passed to the back-end distributed PGO for further joint optimization.

$\mathbf{Pose \ Graph \ Optimization.}$
The PGO submodule uses the odometry measurements and the loop closures measurements (intra-robot and inter-robot) to estimate the trajectories of the robots by solving a maximum likelihood problem:
\begin{equation}
x^{*} = \underset{x}{\arg\max} \prod \phi(x).
\end{equation}
In our implementation, the DGS method \cite{Choudhary16icra} is adopted, in which each robot solves its pose graph using a distributed two-stage Gauss-Seidel method. 
Executing periodically and synchronously within each robot, it estimates the rotations of the trajectories of the robots and then estimates the full pose with the optimized rotations.
During this process, the necessary rotation and pose estimates will be transmitted to the specified robot depending on the optimization order of all connected robots.
With this approach, the robots in the swarm can exchange minimal information to avoid complex bookkeeping and double counting and reach a consensus on the optimal trajectory estimate.
The collaborative localization process is summarized in Algorithm \ref{alg1}.

%% file: experiment.tex
\section{EXPERIMENT}
\label{sec:experiment}
This section showcases the performance, accuracy, and communication of DCL-SLAM evaluated in several scenarios and provides experimental results compared to other state-of-art C-SLAM and single-robot systems. 

\subsection{Experimental Setup}
\label{setup}
\subsubsection{Single-robot SLAM and C-SLAM systems}
The proposed DCL-SLAM system results from the combination of many frameworks and libraries.
The system uses the robot operating system (ROS) to interface with the attached sensors and handles information of different modules exchanged between robots.
All methods are implemented in C++ and executed using ROS in Ubuntu 18.04 LTS.
To show the extensibility of our framework, we integrate existing open-source LiDAR odometry methods, including LIO-SAM and FAST-LIO2.
Moreover, we adopt the C++ implementation of LiDAR-Iris with default parameters (the descriptor size is 80$\times$360) and exchange descriptors among the robotic team. Furthermore, the distributed closure candidates are found by searching the row key of the descriptor with the k-nearest neighbor algorithm implemented in the libnabo library.
As for back-end optimization, we refer to distributed estimation method in \cite{Choudhary16icra} to implement the distributed PGO module using the Georgia Tech Smoothing and Mapping (GTSAM) library.


\subsubsection{Hardware Setup}
\begin{figure}[ht]
	\centering
	\subfloat[]
	{
		\label{fig:setup:a} 
		\raisebox{0.1\height}{\includegraphics[width=0.35\linewidth]{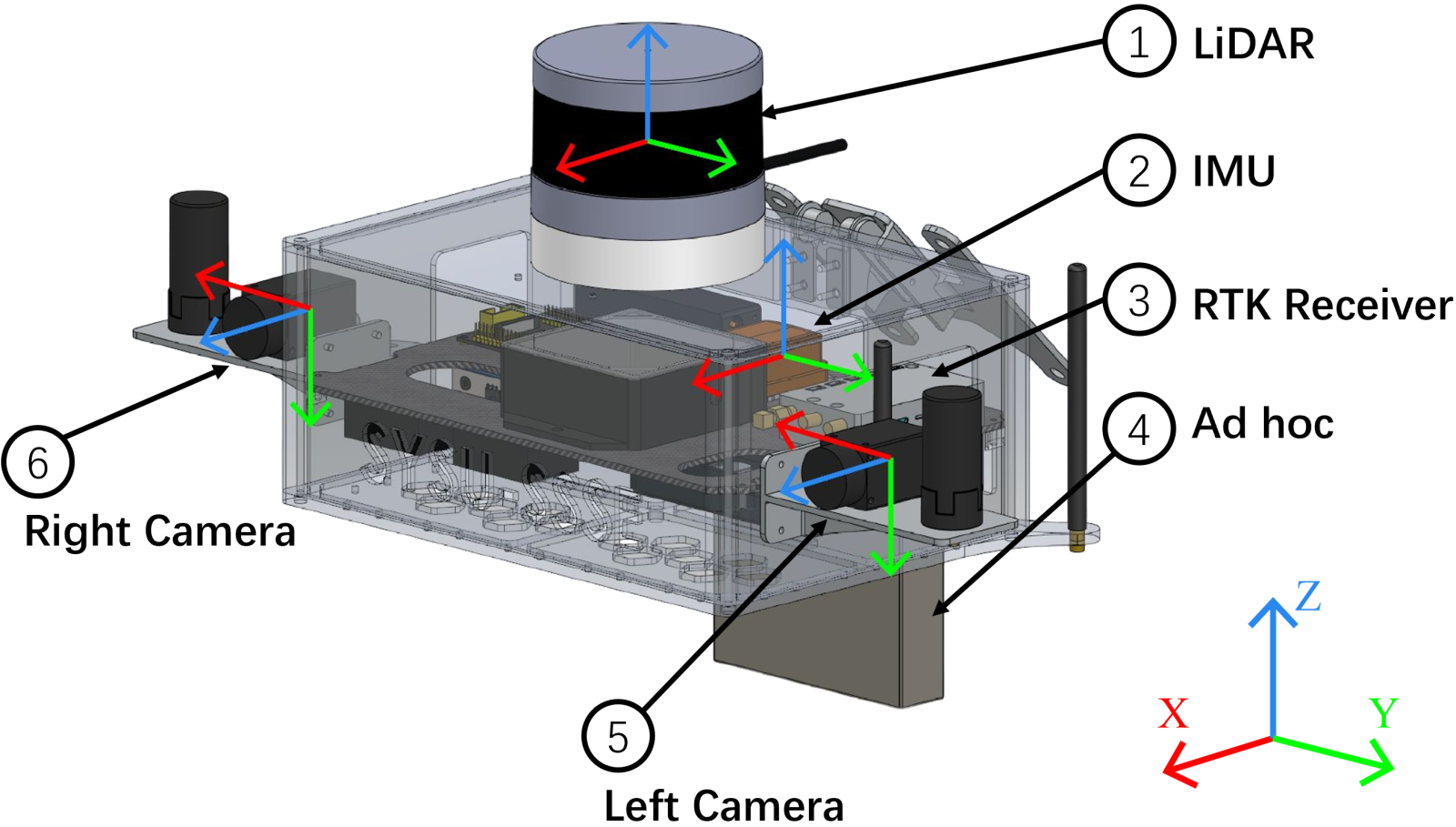}}
	}
	\subfloat[]
	{
		\label{fig:setup:b} 
		\includegraphics[width=0.57\linewidth]{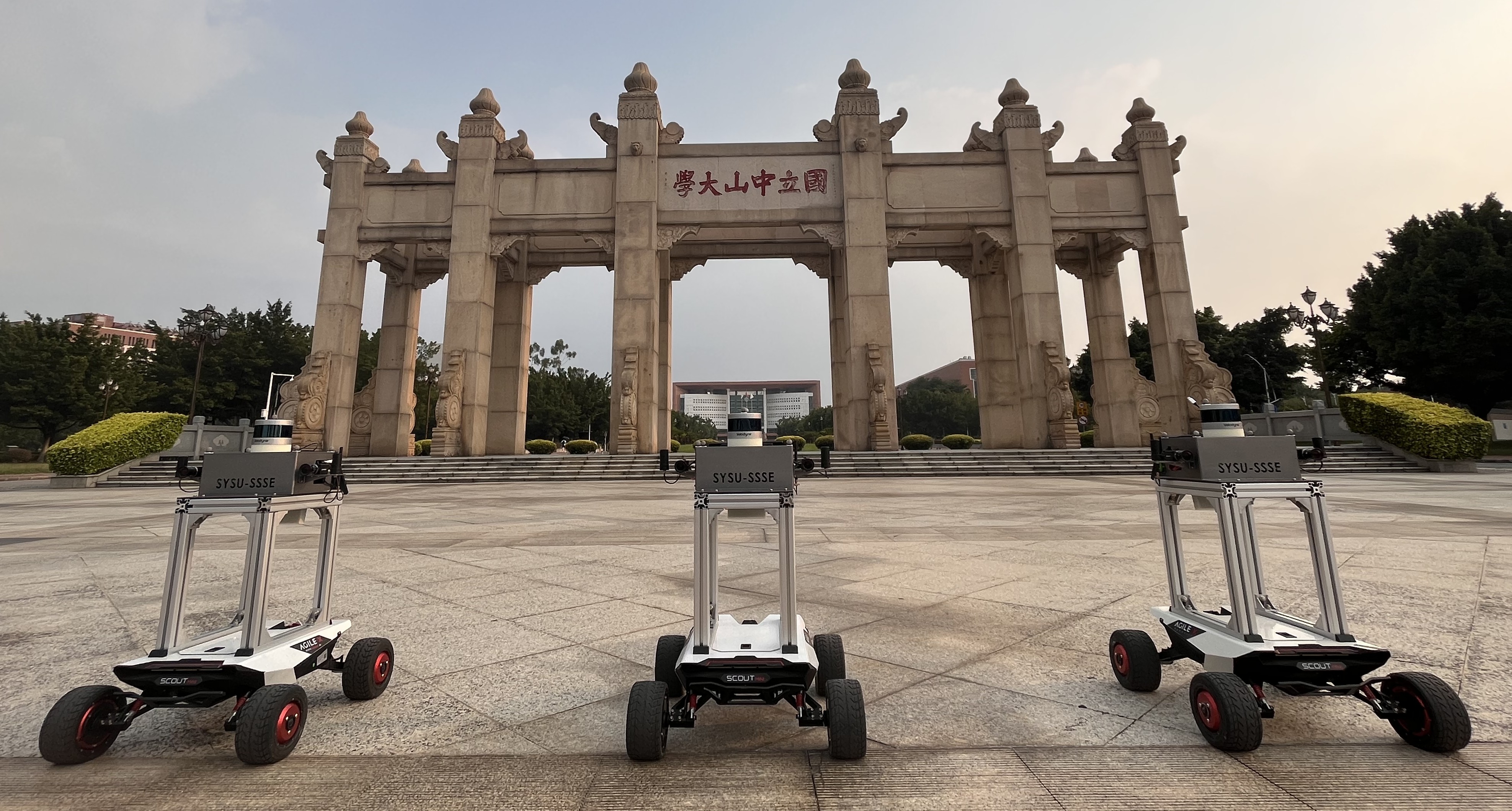}
	}
	\caption{(a) The sensor configuration for our own datasets collected at Sun Yat-Sen University's eastern campus. Note that the attached cameras doesn't use in the following test scenes. (b) Robotic swarm of Agilex SCOUT MINI. }
	\label{fig:hardwaresetup}
\end{figure}
As shown in Fig. \ref{fig:setup:a}, the UGVs used in our tests are Agilex SCOUT MINI, equipped with a Velodyne VLP-16 Puck LiDAR, an Xsens MTi-30 IMU, two HikRobot MV-CS050-10GC cameras,  and an onboard computer NUC11TNKv7 (Intel i7-1165G7 CPU, 16GB DDR4 RAM).
The UGV relies on communication radio NexFi MF1400, creating an ad-hoc wireless network to provide peer-to-peer communication between robots.

Time synchronization and calibration is essential to achieve the high performance of multi-sensor localization and mapping.
For time synchronization, an FPGA board (Altera EP4CE10) is used as the external trigger to periodically generates a pulse to synchronize the LiDAR, cameras, and IMU.
The LiDAR is configured to have a 10 Hz update rate and refreshes its internal timer after receiving the 1Hz trigger pulse.
At the same time, the cameras/IMU collects and returns the data immediately after receiving a 10Hz/100Hz trigger pulse.
We use standard chessboard detection to estimate the camera's intrinsic parameters for intrinsic calibration, while the IMU intrinsic calibration was conducted using the Kalibr \cite{maye2013self} toolbox. Moreover, LiDAR's proprietary calibration model is applied to correct the data during capture directly.
The stereo camera joint and LiDAR-camera joint calibration \cite{zhang2000cameracalibration, zhou2018extrinsiccalibration} is performed for extrinsic calibration. Moreover, the camera-IMU joint calibration is conducted using the Allan standard deviation \cite{furgale2013spatialcalibration}.
Note that the cameras participate in the multi-sensor calibration but does not use in the experiments.

\subsubsection{Datasets}
\begin{table}[tbp]
	\centering
	\caption{Details of Public and Our Collected Datasets}
	\label{tab:dataset infomation}
	\begin{center}
		\tabcolsep=0.13cm
		\renewcommand\arraystretch{1.15}
		\begin{tabular}{cccccc}
			\toprule
			\multicolumn{2}{c}{\multirow{2}{*}{$\mathbf{Datasets}$}} & \multirow{2}{*}{$\mathbf{Environment}$} & \multicolumn{3}{c}{$\mathbf{Trajectory\ Length}$ {[}m{]}} \\
			&  & \multicolumn{1}{c}{} & Robot $\alpha$ & Robot $\beta$ & Robot $\gamma$ \\
			\midrule
			\multirow{3}{*}{\rotatebox{90}{KITTI}} & \textit{Seq. 05} & urban & 424.9 & 599.9 & 1066.7\\
			& \textit{Seq. 08} & urban & 1252.8 & 1991.6 & -\\
			& \textit{Seq. 09} & urban & 557.4 & 733.2 & 544.4\\
			\midrule
			\multirow{7}{*}{\rotatebox{90}{SYSU's Campus}} & Square\_1 & campus & 455.6 & 454.4 & 458.2\\
			& \textit{Square\_2} & campus & 250.6 & 246.4 & -\\
			& \textit{Library} & campus & 507.6 & 517.2 & 498.9\\
			& \textit{Playground} & campus & 407.7 & 425.6 & 445.5 \\
			& \textit{Dormitory} & campus & 727.0 & 719.3 & 721.9 \\
			& \textit{College} & campus & 920.5 & 995.9 & 1072.3 \\
			& \textit{Teaching\_Building} & indoor & 617.2 & 734.4 & 643.4\\
			\bottomrule
		\end{tabular}
	\end{center}
\end{table}

All experimental results in this paper are obtained by comparing performance on three KITTI odometry sequences (HDL-64) and seven collected on SYSU's eastern campus (VLP-16).
For KITTI odometry sequences, we especially adopt the sequences 05, 08, and 09 among 11 sequences, as they have a variety of loop closure types, including the same and opposite viewpoints.
Furthermore, we split sequences 05 and 09 into three parts and sequence 08 into two to simulate multi-robot scenarios.
The sequences from KITTI provide GNSS ground-truth data for experimental evaluation.
In addition, six outdoor sequences and one indoor sequence are collected with a robotic team in different scenarios at Sun Yat-Sen University's eastern campus, as shown in Fig. \ref{fig:setup:b}.
The indoor sequence is an online experiment specifically for outliers rejection.
Real-time kinematic (RTK) provide the ground-truth data in the outdoor sequences. 
Table \ref{tab:dataset infomation} recaps the main characteristics for all sequences used in the following experiments.
Aligned all robots with a single alignment obtained by aligning the prior owner robot, we used the evo \cite{grupp2017evo}
toolbox to evaluate and compare the trajectory output of odometry and C-SLAM algorithms for KITTI and our campus sequences.

\begin{figure*}[tbp]
	\centering
	\subfloat[\textit{Seq. 05}]
	{
		\label{fig:kitti05prc} 
		\includegraphics[width=0.3\linewidth]{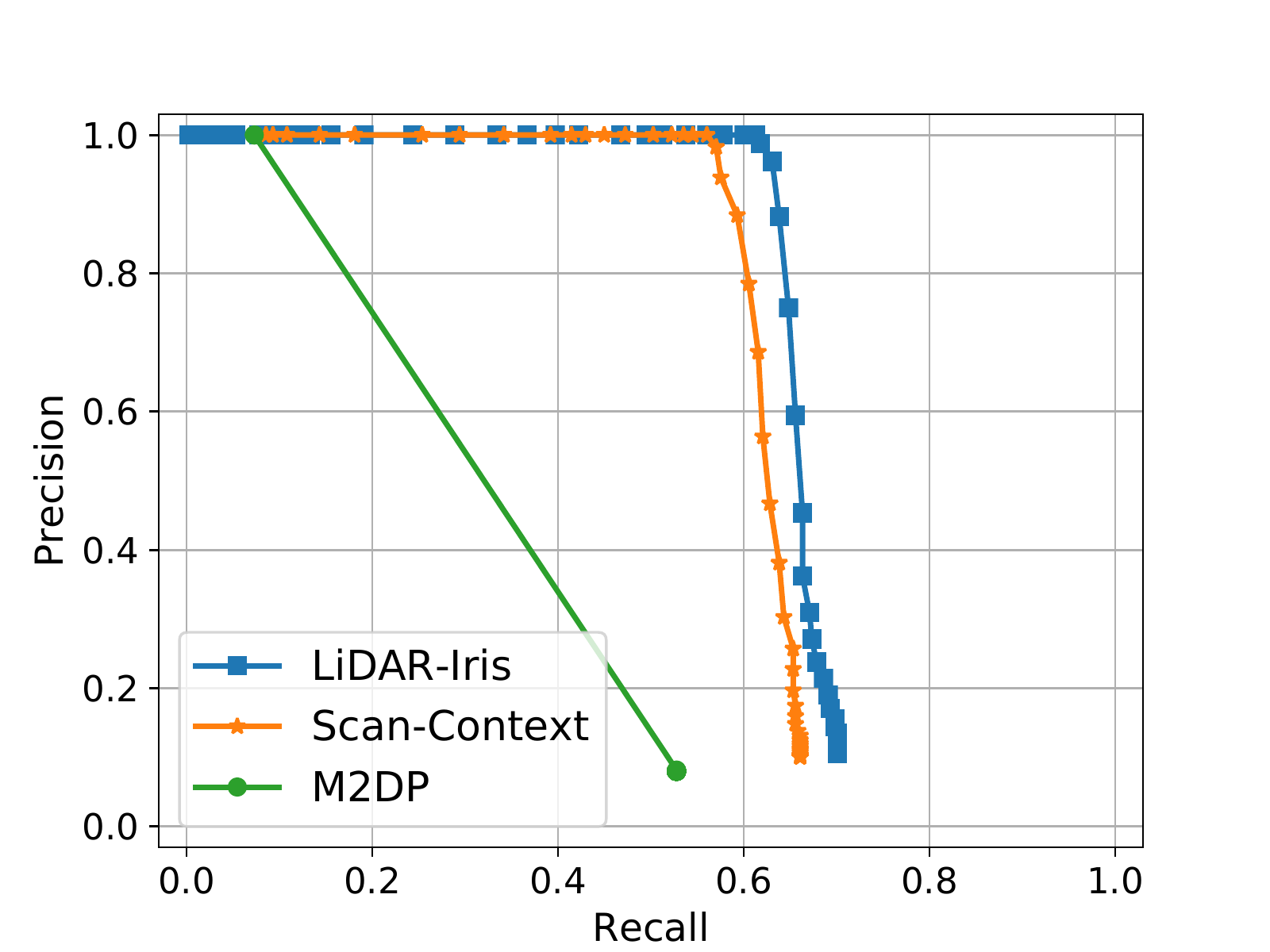}
	}
	\subfloat[\textit{Seq. 08}]
	{
		\label{fig:08} 
		\includegraphics[width=0.3\linewidth]{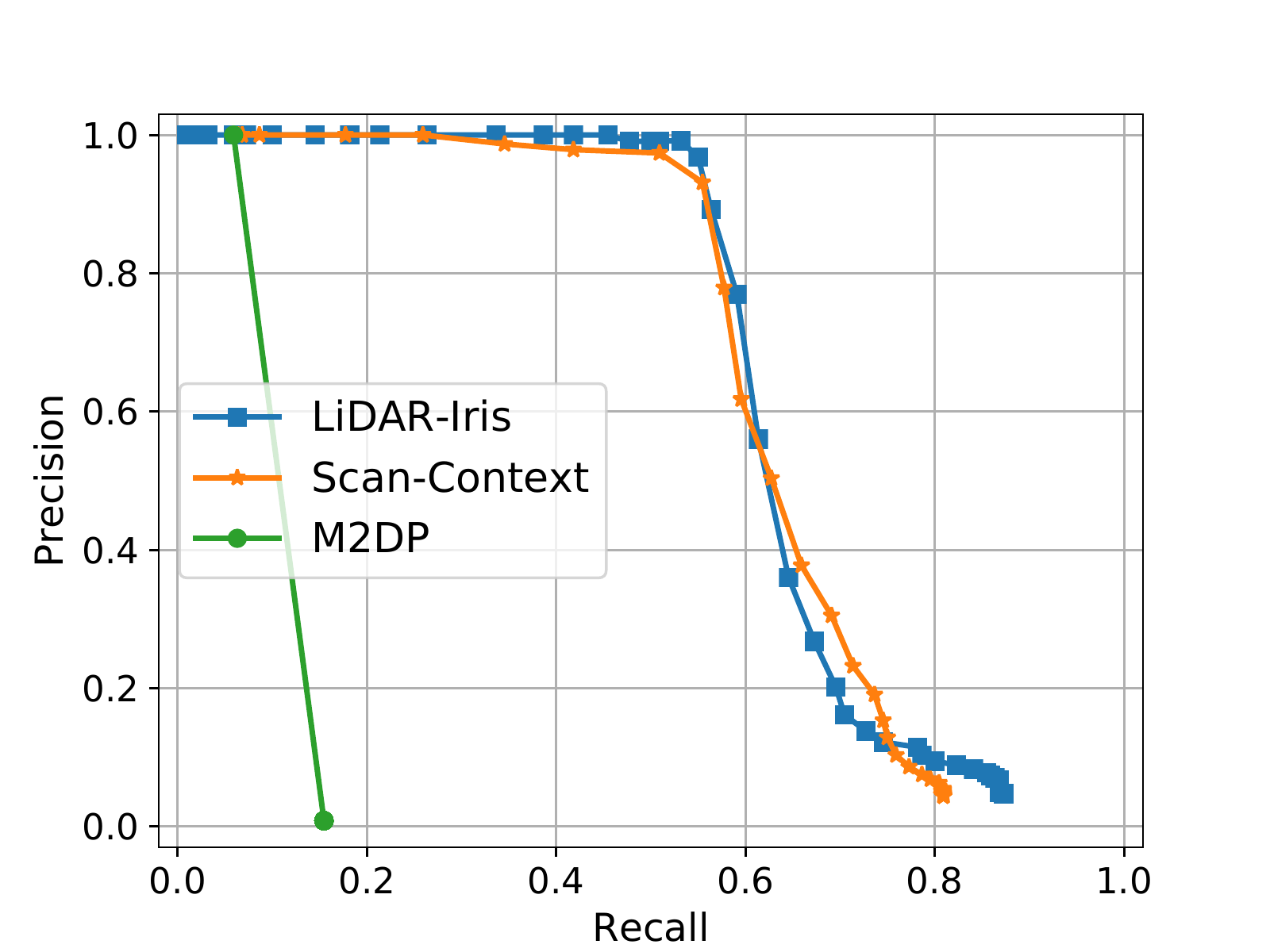}
	}
	\subfloat[\textit{Seq. 09}]
	{
		\label{fig:kitti09prc} 
		\includegraphics[width=0.3\linewidth]{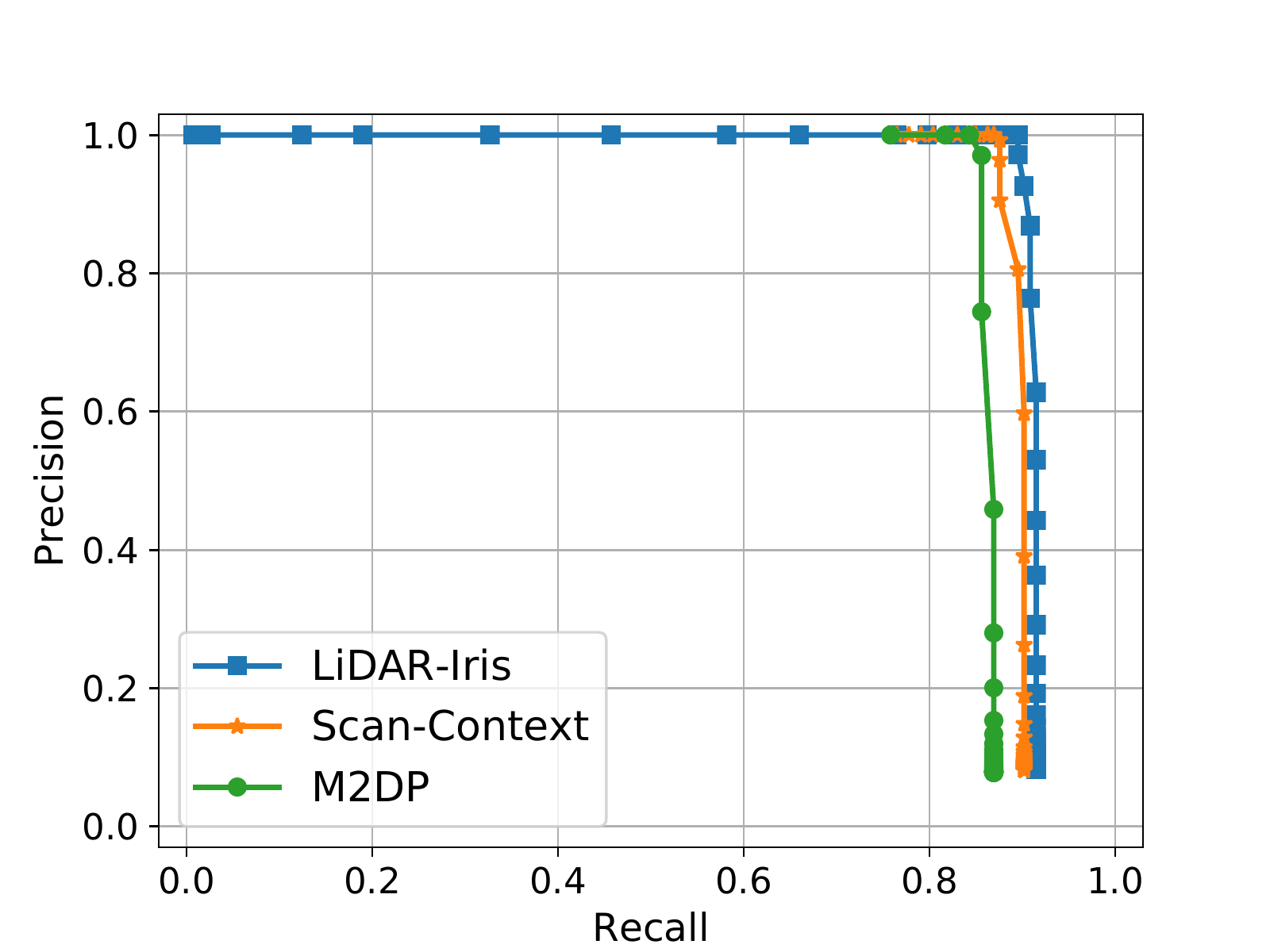}
	}
	\caption{The precision-recall curves of different methods on different KITTI sequences.}
	\label{fig:prc}
\end{figure*}
\begin{figure*}[t]
	\centering
	\subfloat[LIO-SAM]
	{
		\label{fig:lio_sam_kitti05} 
		{\includegraphics[width=0.18\linewidth]{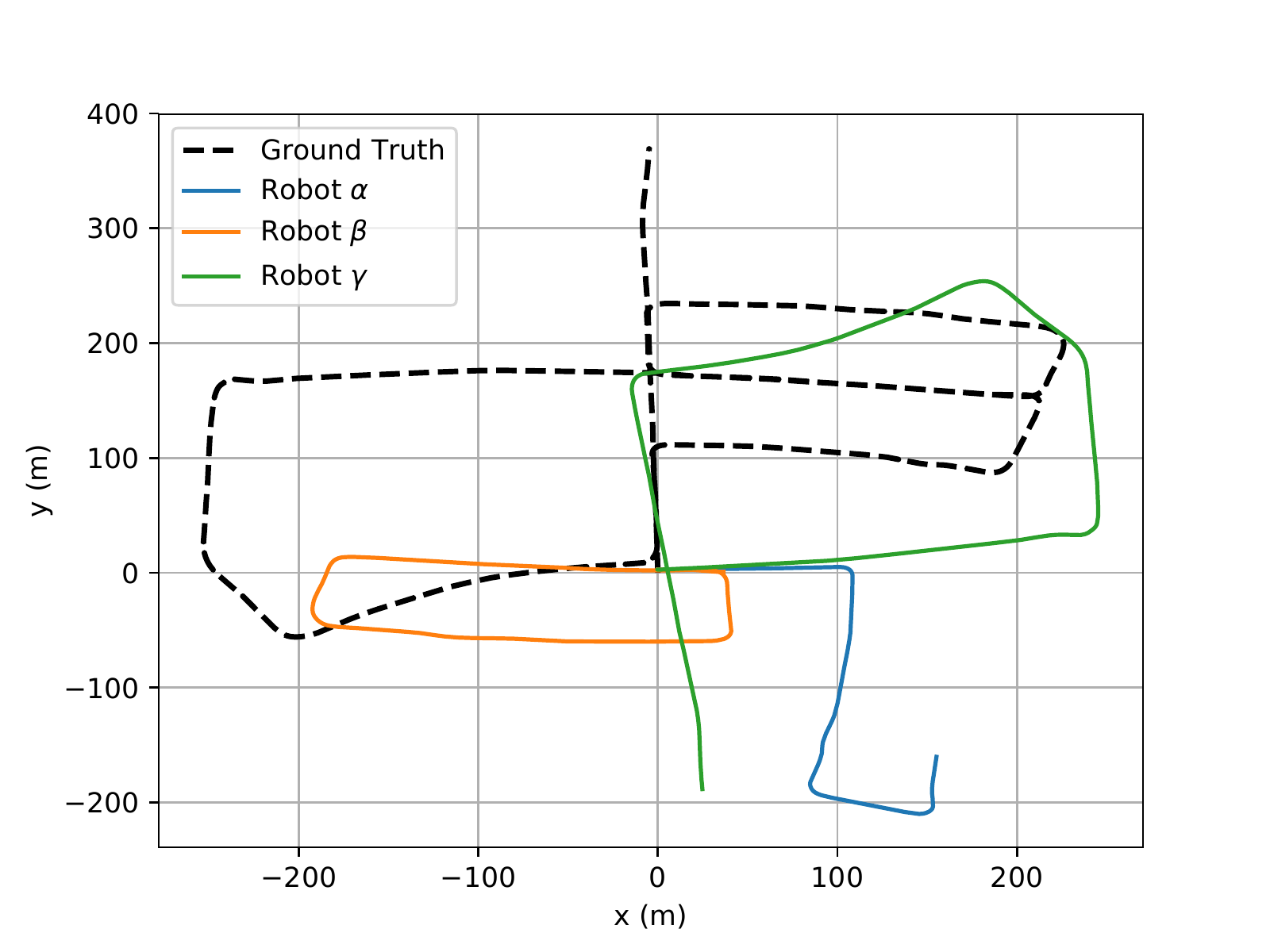}}
	}
	\subfloat[FAST-LIO2]
	{
		\label{fig:fast_lio_kitti05} 
		{\includegraphics[width=0.18\linewidth]{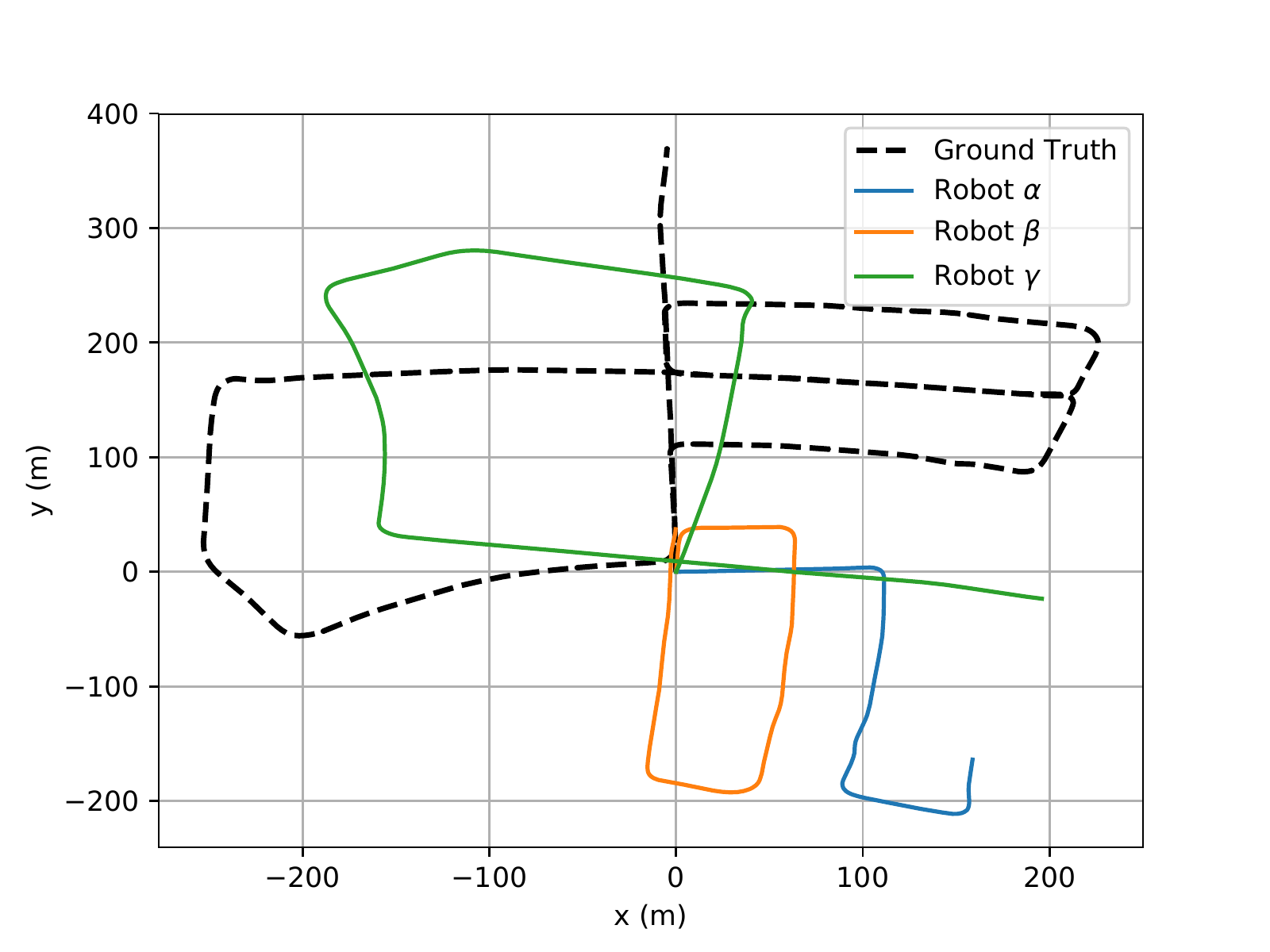}}
	}
	\subfloat[DiSCo-SLAM]
	{
		\label{fig:disco_slam_kitti05} 
		\includegraphics[width=0.18\linewidth]{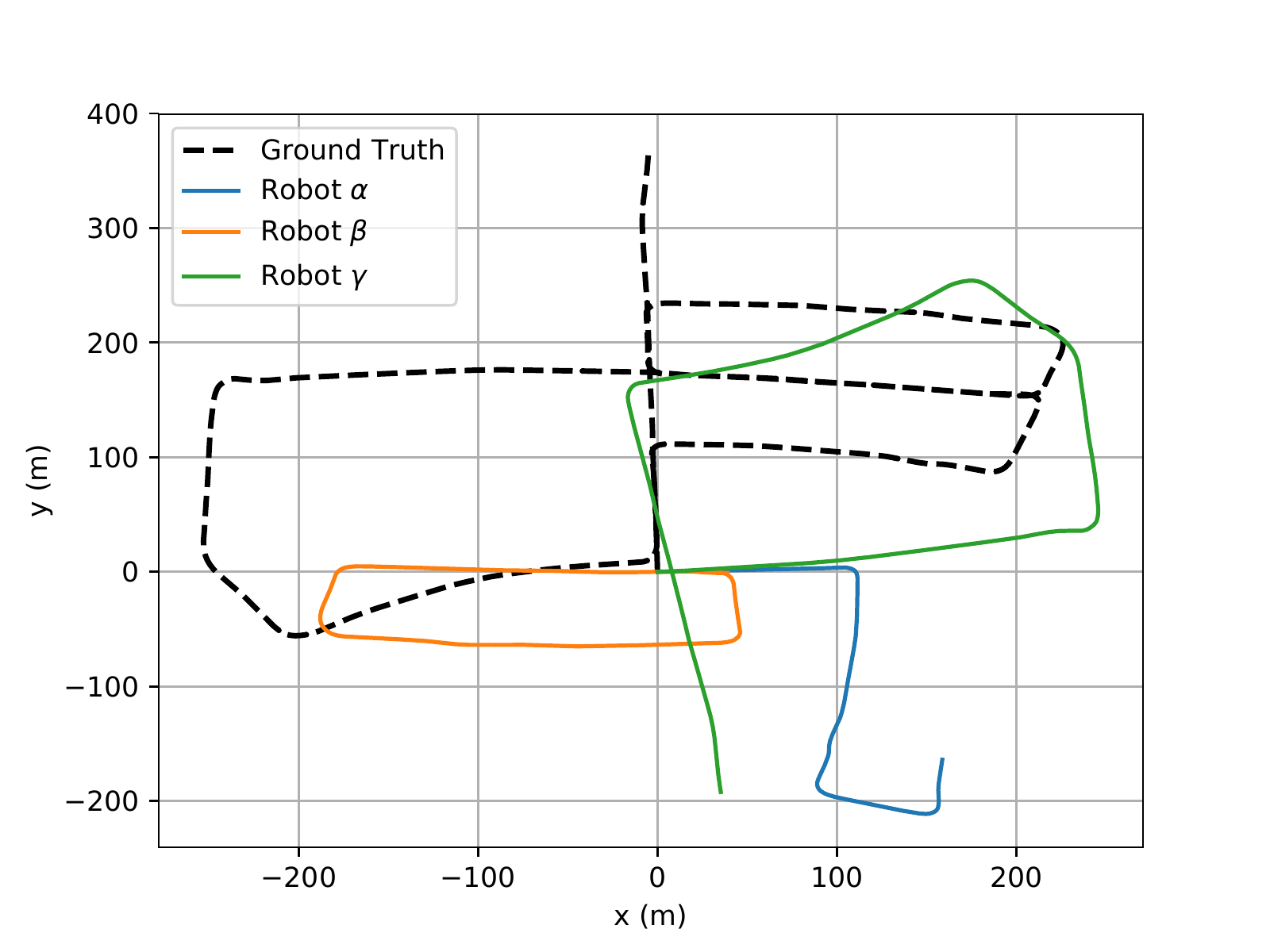}
	}
	\subfloat[DCL-LIO-SAM]
	{
		\label{fig:dcl_lio_sam_kitti05} 
		{\includegraphics[width=0.18\linewidth]{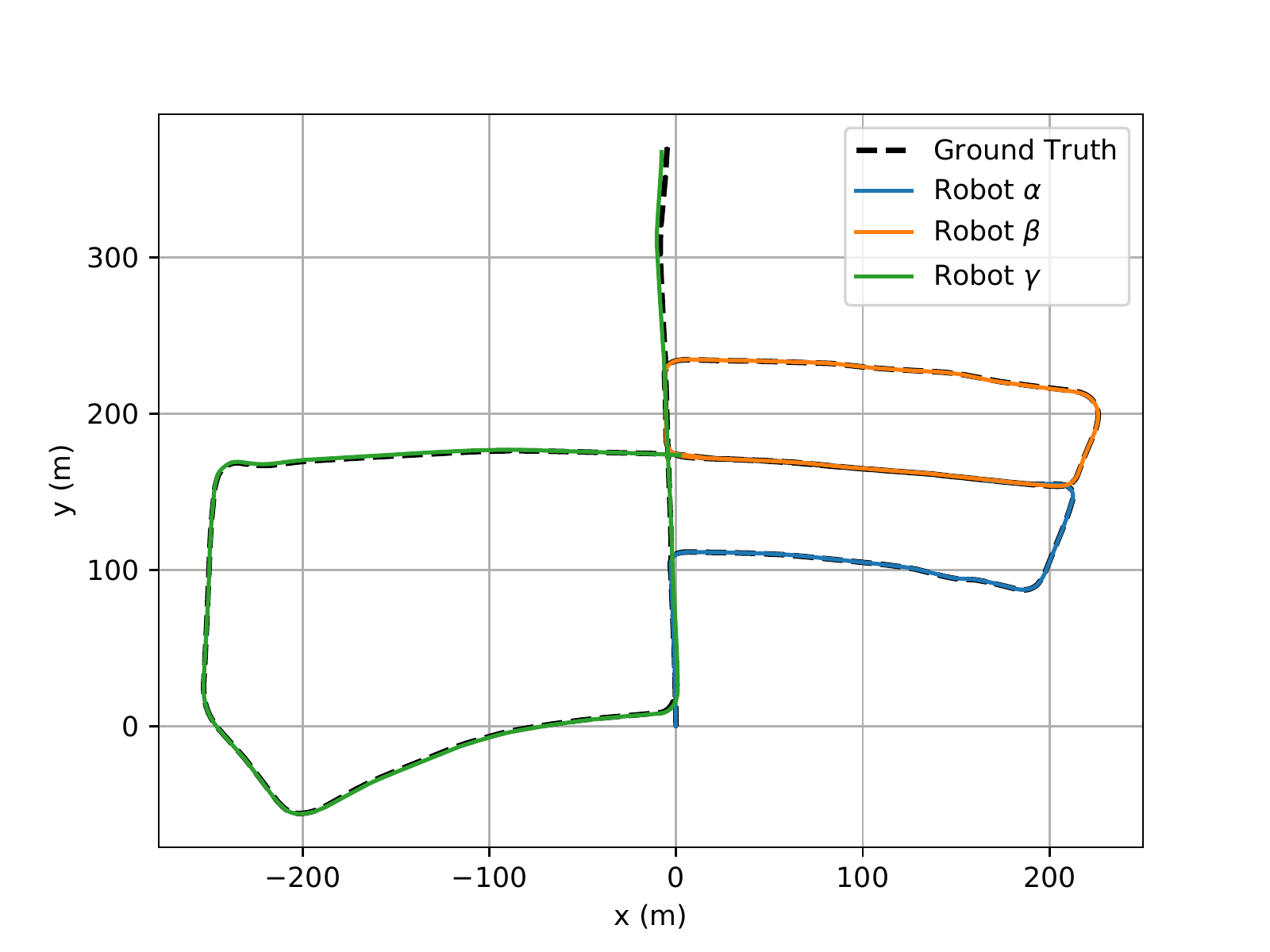}}
	}
	\subfloat[DCL-FAST-LIO2]
	{
		\label{fig:dcl_fast_lio_kitti05} 
		{\includegraphics[width=0.18\linewidth]{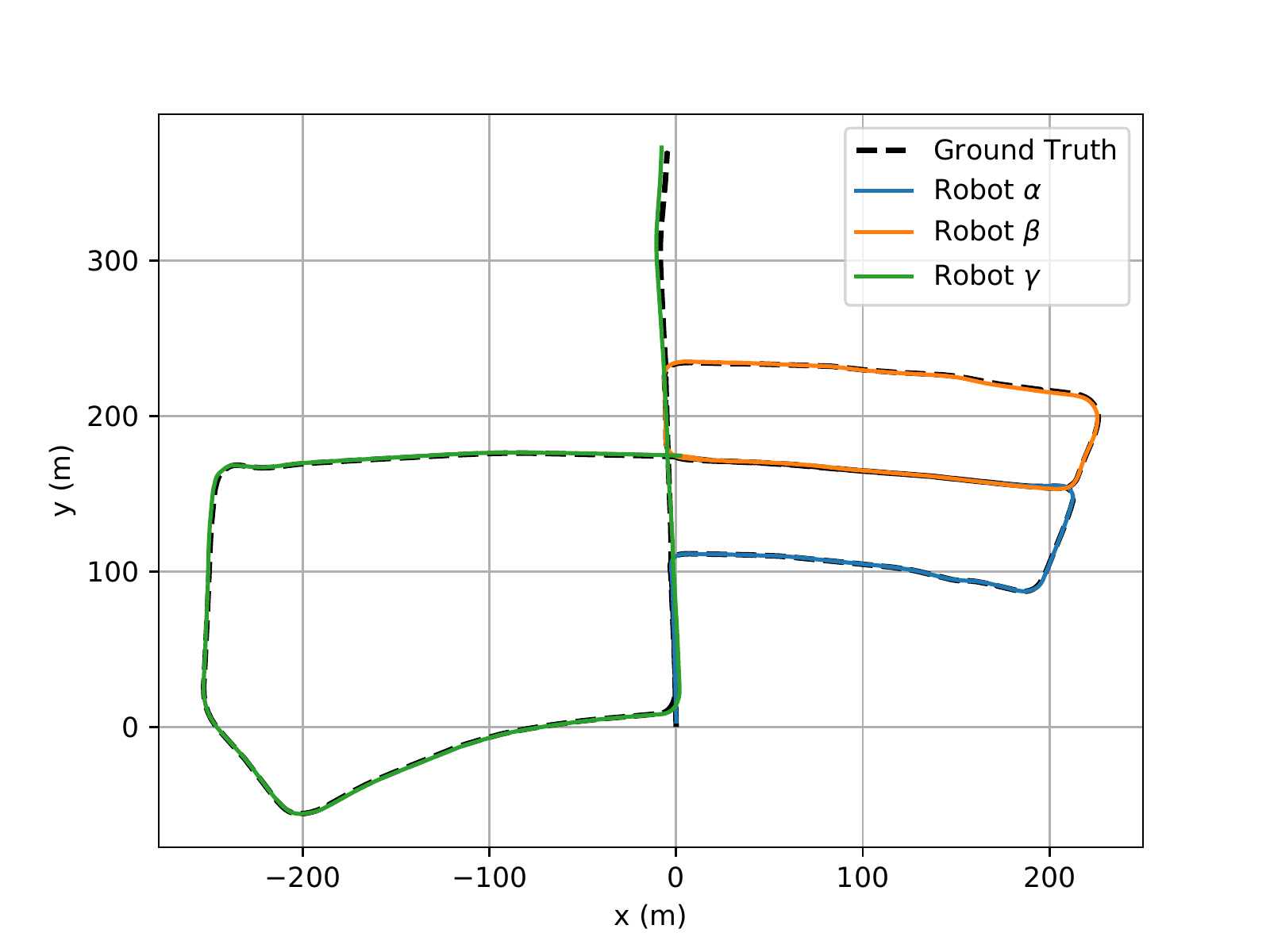}}
	}\par
	\subfloat[LIO-SAM]
	{
		\label{fig:lio_sam_library} 
		{\includegraphics[width=0.18\linewidth]{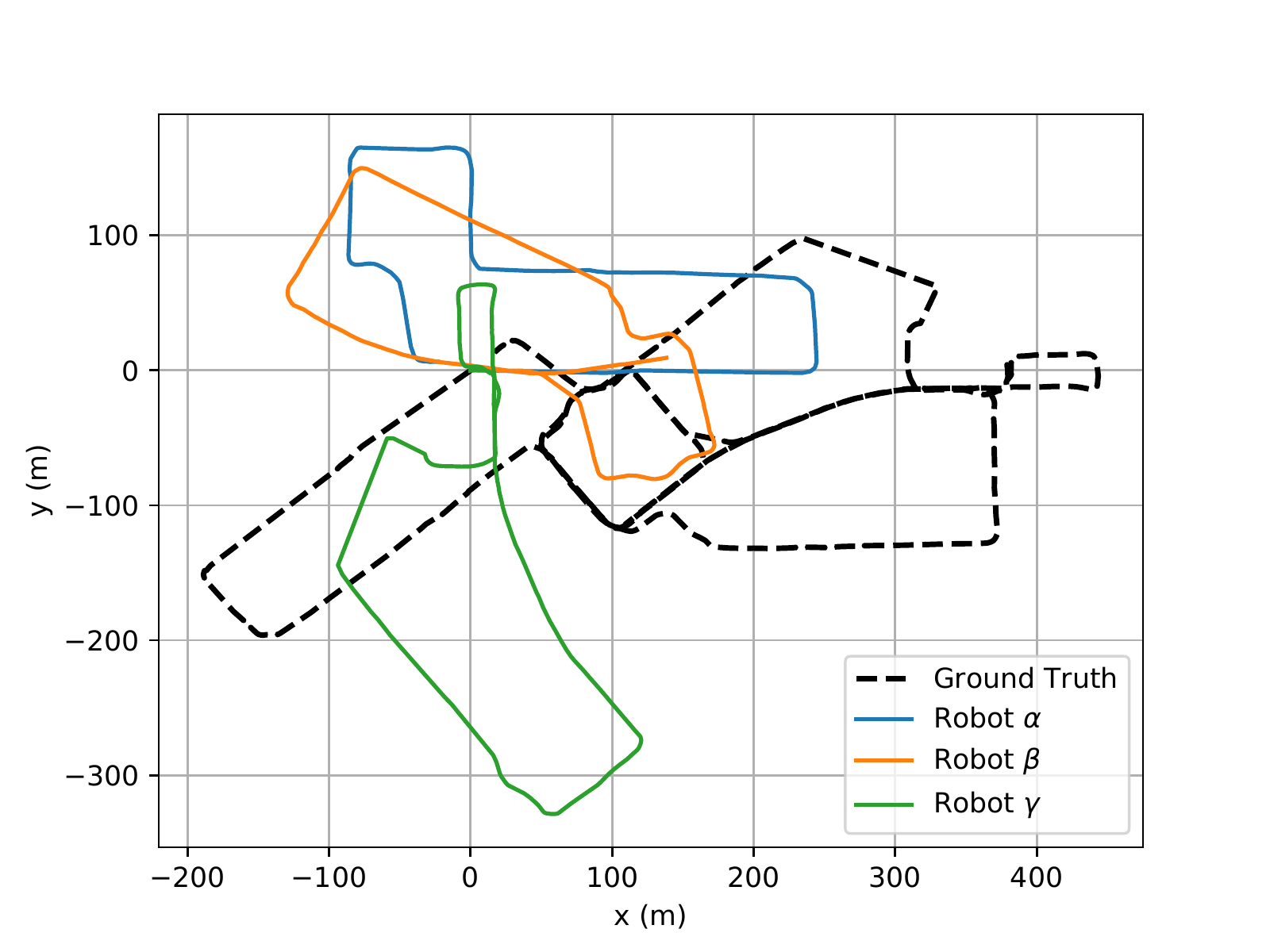}}
	}
	\subfloat[FAST-LIO2]
	{
		\label{fig:fast_lio_library} 
		{\includegraphics[width=0.18\linewidth]{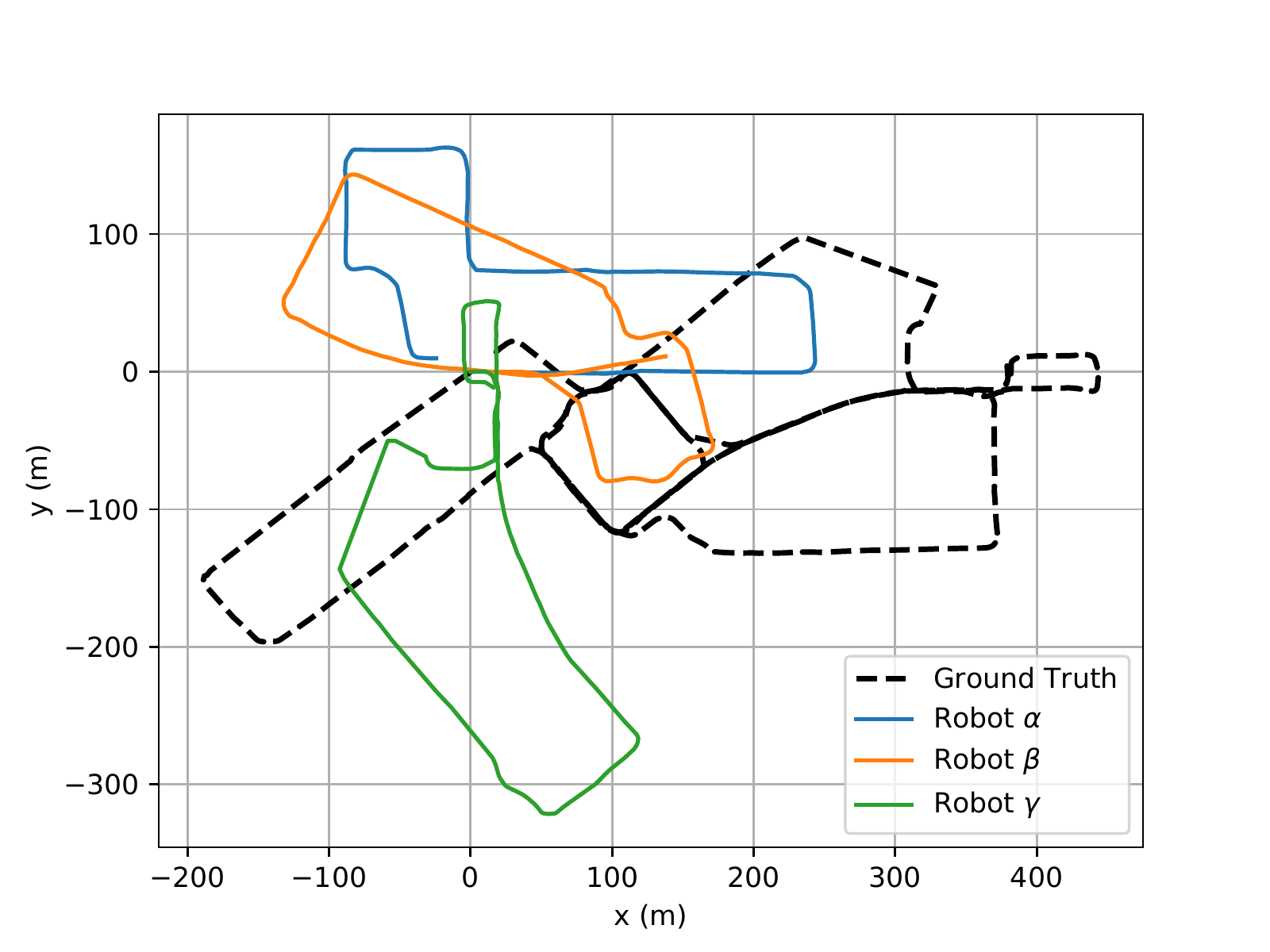}}
	}
	\subfloat[DiSCo-SLAM]
	{
		\label{fig:disco_slam_library} 
		\includegraphics[width=0.18\linewidth]{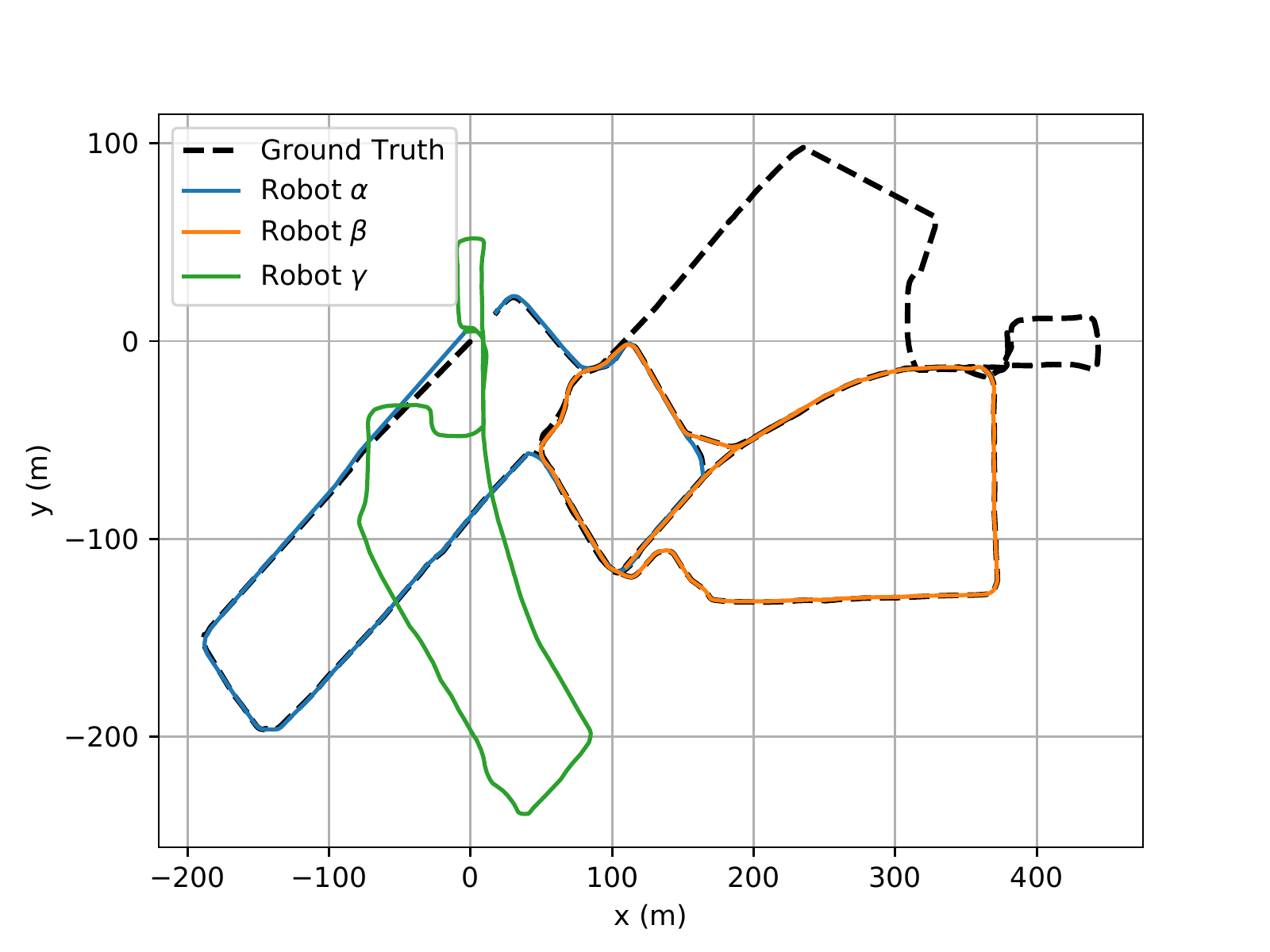}
	}
	\subfloat[DCL-LIO-SAM]
	{
		\label{fig:dcl_lio_sam_library} 
		{\includegraphics[width=0.18\linewidth]{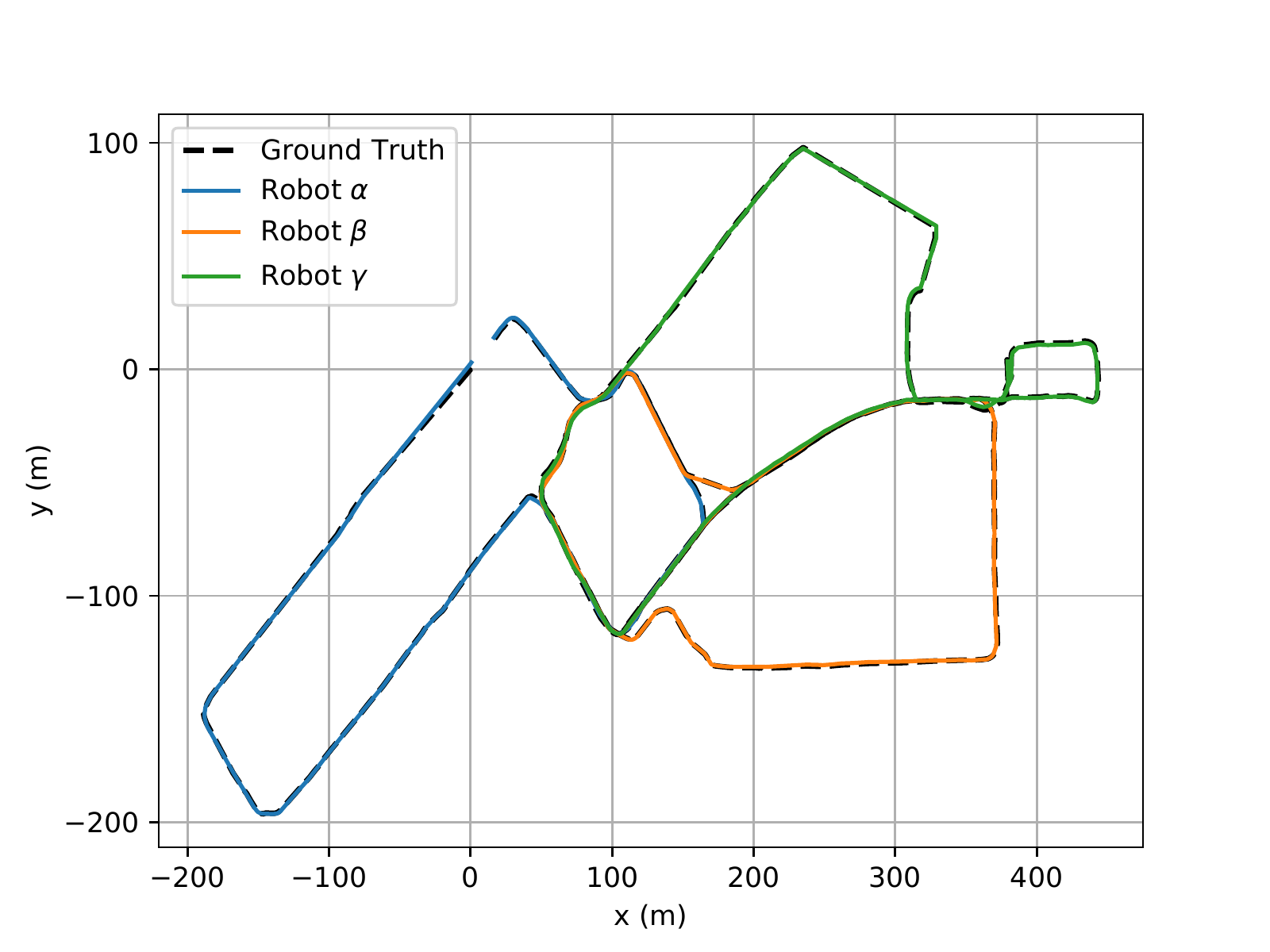}}
	}
	\subfloat[DCL-FAST-LIO2]
	{
		\label{fig:dcl_fast_lio_library} 
		{\includegraphics[width=0.18\linewidth]{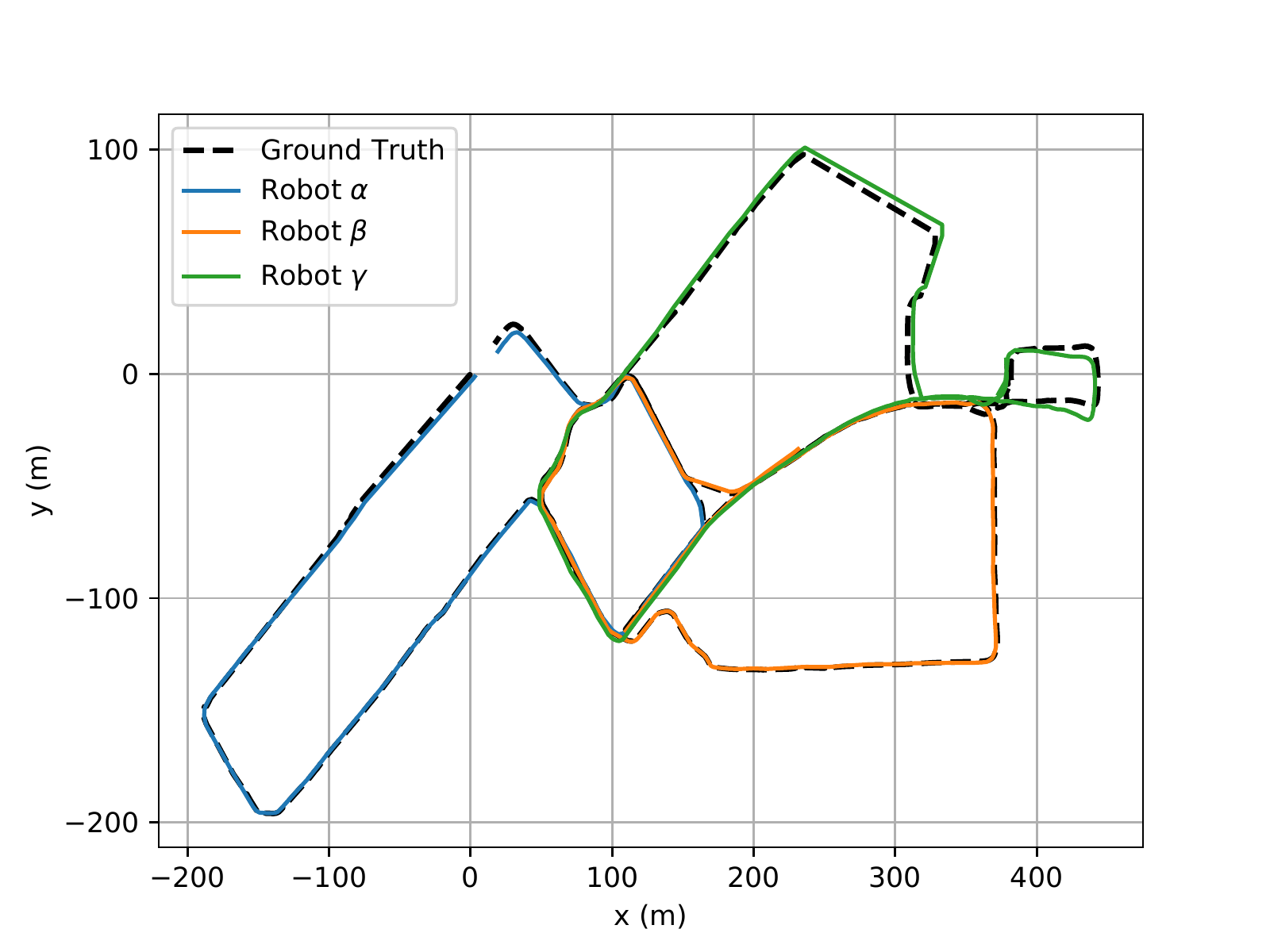}}
	}
	\caption{The trajectories result over the KITTI \textit{Seq. 05} (from (a) to (c)) and our \textit{College} (from (d) to (f)), with different method.}
	\label{fig:trajectoryandmap}
\end{figure*}

\subsection{Performance of Distributed Loop Closure}
To compare the end-to-end distributed loop closure detection performance and obtain the best threshold of matching distance, we used the original laser scan from the KITTI sequences to benchmark all global descriptor methods and assumed that all robots start simultaneously.
The closest candidate produced by matching the current query with the history database is regarded as an inter-robot loop closure candidate and further verified against the ground truth.
If the ground truth distance between the current query and the closest match is less than 4m, the loop closure candidate is regarded as a successful loop closure.
Note that the 4m-distance is set as default according to \cite{kim2018_scancontext}, and default parameters are set for LiDAR-Iris, Scan-Context, and M2DP.

The precision-recall (PR) curves of different methods are shown in Fig. \ref{fig:prc}. Precision is the percentage of successful loop closures, and recall rate is the reported loop closures against total loop closures.
For reaching a 100\% precision, the maximum recall rate is 0.61, 0.45 and 0.90 for LiDAR-Iris on \textit{Seq. 05}, \textit{Seq. 08} and \textit{Seq. 09}, while 0.073, 0.059 and 0.84 for M2DP, and 0.56, 0.26 and 0.87 for Scan-Context.
Furthermore, these descriptors are applied to detect the loop closure in the SLAM system with the same parameter for verification and optimization. The distance thresholds are set to 0.3, 0.35, and 0.3 for LiDAR-Iris, Scan-Context, and M2DP according to the PR curves.
The average ATE of the trajectories is 0.59, 4.90, and 1.11 for LiDAR-Iri on \textit{Seq. 05}, \textit{Seq. 08} and \textit{Seq. 09}, while 0.76, 5.01, and 1.18 for Scan-Context, and 0.80, 9.96, and 2.23 for M2DP.
M2DP only reported high precision with a lower recall rate and high accuracy on \textit{Seq. 09}, which shows that M2DP can detect simple inter-loops correctly.
Compared to M2DP, Scan-Context can achieve promising results on all three sequences.
In contrast, LiDAR-Iris demonstrates very competitive performance on all three sequences and achieves the best performance, which shows its advanced performance in detecting various inter-robot loop closures.
Regarding computational complexity, the average matching time of LiDAR-Iris on \textit{Seq. 08} is 24.57ns, the time of Scan-Context is 6.71ns, and the time of M2DP is 0.01ns.

\subsection{System Evaluation}
\begin{table*}[t]
	\centering
	\caption{ATE W.r.t RTK of the SYSU's Eastern Campus Datasets}
	\label{tab:outdoor dataset evaluation}
	\begin{center}
		\tabcolsep=0.13cm
		\renewcommand\arraystretch{1.15}
		\begin{tabular}{c|ccc|cc|ccc|ccc|ccc|ccc}
			\toprule
			\multicolumn{1}{c}{\multirow{2}{*}{\diagbox{Configure}{Datasets}}} & \multicolumn{3}{c}{\textit{Square\_1}} & \multicolumn{2}{c}{\textit{Square\_2}} & \multicolumn{3}{c}{\textit{Library}} & \multicolumn{3}{c}{\textit{Playground}} & \multicolumn{3}{c}{\textit{Dormitory}} & \multicolumn{3}{c}{\textit{College}} \\
			\multicolumn{1}{c}{} & \tiny{Robot $\alpha$} & \tiny{Robot $\beta$} &\tiny{Robot $\gamma$} &\tiny{Robot $\alpha$} & \tiny{Robot $\beta$} & \tiny{Robot $\alpha$} & \tiny{Robot $\beta$} &\tiny{Robot $\gamma$} & \tiny{Robot $\alpha$} & \tiny{Robot $\beta$} &\tiny{Robot $\gamma$}& \tiny{Robot $\alpha$} & \tiny{Robot $\beta$} &\tiny{Robot $\gamma$} &\tiny{Robot $\alpha$} & \tiny{Robot $\beta$} &\tiny{Robot $\gamma$}\\
			
			\midrule
			
			DiSCo-SLAM & \multicolumn{3}{c|}{Failed}& \multicolumn{2}{c|}{Failed} & 0.74&1.33&1.27 & $\mathbf{0.27}$&$\mathbf{0.37}$&0.38 & \multicolumn{3}{c|}{Failed} & \multicolumn{3}{c}{Failed}\\
			
			only LIO-SAM &  $\mathbf{1.23}$&0.92&0.69 & $\mathbf{0.46}$&0.27 & 1.03&1.74&1.30 & 0.35&0.72&0.45 & 0.82&1.71&0.70 & 2.35&4.58&$\mathbf{1.41}$\\
			
			our DCL-LIO-SAM &  $\mathbf{1.23}$&$\mathbf{0.86}$&$\mathbf{0.66}$ & 0.50&$\mathbf{0.26}$ & $\mathbf{0.58}$&$\mathbf{1.26}$&1.17 & 0.33&0.40&$\mathbf{0.32}$ & $\mathbf{0.52}$&$\mathbf{1.37}$&$\mathbf{0.51}$ & $\mathbf{1.51}$&$\mathbf{1.48}$&1.87 \\
			
			only FAST-LIO2 & 1.58&2.87&0.72 & 1.47&0.60 & 1.53&2.29&1.10 & 0.48&0.76&0.77 &  9.30&6.28&2.39 & 3.63&5.69&12.88\\
			
			our DCL-FAST-LIO2 & 1.56&2.87&0.72 & 1.45&0.60 & 1.30&1.38&$\mathbf{1.07}$ & 0.58&0.68&0.40 & 6.91&5.64&2.33 & 6.13&3.27&3.30\\
			
			
			\bottomrule
		\end{tabular}
	\end{center}
\end{table*}

\begin{table}[!t]
	\centering
	\caption{Estimation Accuracy of the KITTI Odometry Sequences}
	\label{tab:kitti_evaluation}
	\begin{center}
		\tabcolsep=0.1cm
		\renewcommand\arraystretch{1.15}
		\begin{tabular}{cccccccc}
			\toprule
			\multicolumn{1}{c}{\multirow{2}{*}{$\mathbf{Seq.}$}} & \multicolumn{1}{c}{\multirow{2}{*}{$\mathbf{Configure}$}} & \multicolumn{3}{c}{$\mathbf{ATE}$ [m]} & \multicolumn{3}{c}{$\mathbf{ARE}$ [deg]} \\ 
			\multicolumn{1}{c}{} & \multicolumn{1}{c}{} & \tiny{Robot $\alpha$} & \tiny{Robot $\beta$} &\tiny{Robot $\gamma$} &\tiny{Robot $\alpha$} & \tiny{Robot $\beta$} &\tiny{Robot $\gamma$}\\
			\midrule
			
			\multicolumn{1}{c}{\multirow{5}{*}{\textit{Seq. 09}}} & DiSCo-SLAM & \multicolumn{3}{c}{Failed} & \multicolumn{3}{c}{Failed} \\
			\multicolumn{1}{c}{} & only LIO-SAM & 1.04&1.70&0.90 & 1.50&2.82&2.28 \\
			\multicolumn{1}{c}{} & our DCL-LIO-SAM & 1.07&$\mathbf{1.40}$&0.85 & $\mathbf{1.46}$&$\mathbf{2.64}$&2.20 \\
			\multicolumn{1}{c}{} & only FAST-LIO2 & 1.08&4.14&0.77 & 2.91&15.51&1.97 \\
			\multicolumn{1}{c}{} & our DCL-FAST-LIO2 & $\mathbf{0.95}$&2.46&$\mathbf{0.70}$ & 2.83&8.77&$\mathbf{1.73}$ \\
			\midrule
			
			\multicolumn{1}{c}{\multirow{5}{*}{\textit{Seq. 08}}} & DiSCo-SLAM & 5.16&5.73&- & 6.12&$\mathbf{5.47}$&- \\
			\multicolumn{1}{c}{} & only LIO-SAM & 5.57&5.53&- & 6.46&5.71&- \\
			\multicolumn{1}{c}{} & our DCL-LIO-SAM & 4.92&4.87&- & 6.25&5.69&- \\
			\multicolumn{1}{c}{} & only FAST-LIO2 & 5.29&4.98&- & 5.58&6.25&- \\
			\multicolumn{1}{c}{} & our DCL-FAST-LIO2 & $\mathbf{4.58}$&$\mathbf{4.54}$&- & $\mathbf{5.15}$&6.04&- \\
			\midrule
			
			\multicolumn{1}{c}{\multirow{5}{*}{\textit{Seq. 05}}} & DiSCo-SLAM & \multicolumn{3}{c}{Failed} & \multicolumn{3}{c}{Failed} \\
			\multicolumn{1}{c}{} & only LIO-SAM & 0.30&1.21&2.80 & 1.47&1.48&11.34 \\
			\multicolumn{1}{c}{} & our DCL-LIO-SAM & $\mathbf{0.25}$&$\mathbf{0.44}$&$\mathbf{1.08}$ & $\mathbf{1.43}$&$\mathbf{1.14}$&$\mathbf{1.12}$\\
			\multicolumn{1}{c}{} & only FAST-LIO2 & 0.46&0.90&2.30 & 2.51&2.25&2.25 \\
			\multicolumn{1}{c}{} & our DCL-FAST-LIO2 & 0.55&1.05&1.34 & 1.93&2.00&1.82\\
			\bottomrule
		\end{tabular}
	\end{center}
\end{table}

To evaluate the accuracy and the robustness of the whole SLAM system, we compared the proposed method to DiSCo-SLAM and single-robot SLAM systems on KITTI odometry datasets, and the result ATE and ARE are shown in Tab. \ref{tab:kitti_evaluation}. In addition, Fig. \ref{fig:trajectoryandmap}(a)-(d) presents the trajectories result of different methods on \textit{Seq. 05}.
As illustrated in the table, the performance of DCL-SLAM with different odometry, including LIO-SAM (named DCL-LIO-SAM) and FAST-LIO2 (named DCL-FAST-LIO2), is always better than that of only odometry.
The reason is that DCL-SLAM introduces more measurements between robots for back-end optimization.
The table also shows us that DCL-SLAM successfully merges the sub-maps in \textit{Seq. 05} and \textit{Seq. 09} when DiSCo-SLAM failed.
Moreover, our DCL-SLAM has a smaller ATE, but a larger ARE than DiSCo-SLAM on \textit{Seq. 08}, and reach the best in \textit{Seq. 05} and \textit{Seq. 09}.
These illustrate that DCL-SLAM has higher precision and good robustness.

In order to further evaluate the accuracy and robustness of the proposed system and overcome the reality gap, we deployed UGVs on our campus for field tests and recorded datasets, as shown in section \ref{setup}.
Similar to the above public datasets evaluation, we compared the proposed DCL-SLAM to DiSCo-SLAM and single-robot SLAM systems on recorded campus datasets. The ATE result is shown in Tab \ref{tab:outdoor dataset evaluation}.
Furthermore, the results of the trajectories and reconstructed maps of the UGVs on sequence \textit{Library} performed by DLC-LIO-SAM are presented in Fig. \ref{fig:library}.
In addition, Fig. \ref{fig:trajectoryandmap}(e)-(i) presents the trajectories output of different methods on sequence \textit{College}.
These results show that DiSCo-SLAM behaved well in some sequences but failed to merge all or part of the sub-maps in most sequences, which can be caused by poor-designed loop closures searching and matching modules or the wrong default parameter setting.
As opposed to DiSCo-SLAM, our DCL-LIO-SAM approach demonstrates very competitive performance on most sequences and reaches the best on \textit{Square\_1}, \textit{Library}, \textit{Dormitory} and \textit{College} sequences.
Compared to using odometry only, DCL-SLAM reported similar performance on two \textit{Square} sequences because the system can only merge the sub-maps with simple inter-loops, leaving the challenging ones.
Moreover, DCL-SLAM improves performance by over 20\% on the other four sequences.
Note that FAST-LIO has degradation of the LiDAR odometry accuracy of the z-axis on \textit{Dormitory} and \textit{College} sequences, leading to the larger ATE of trajectory compared to the RTK.
In this case, DLC-FAST-LIO2 cannot properly correct the odometry's drift.
In general, DCL-SLAM has high precision and broad applicability.

\subsection{Field Test with Outlier Rejection}
\begin{figure}[t]
	\centering
	\subfloat[without PCM]
	{
		\label{fig:withoutpcm} 
		{\includegraphics[width=0.46\linewidth]{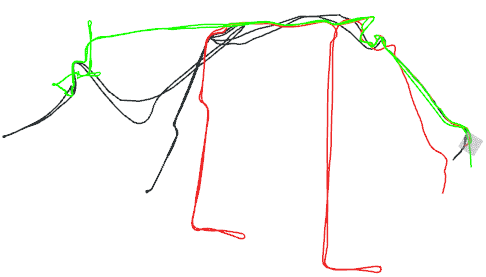}}
	}
	\subfloat[with PCM]
	{
		\label{fig:withpcm} 
		{\includegraphics[width=0.46\linewidth]{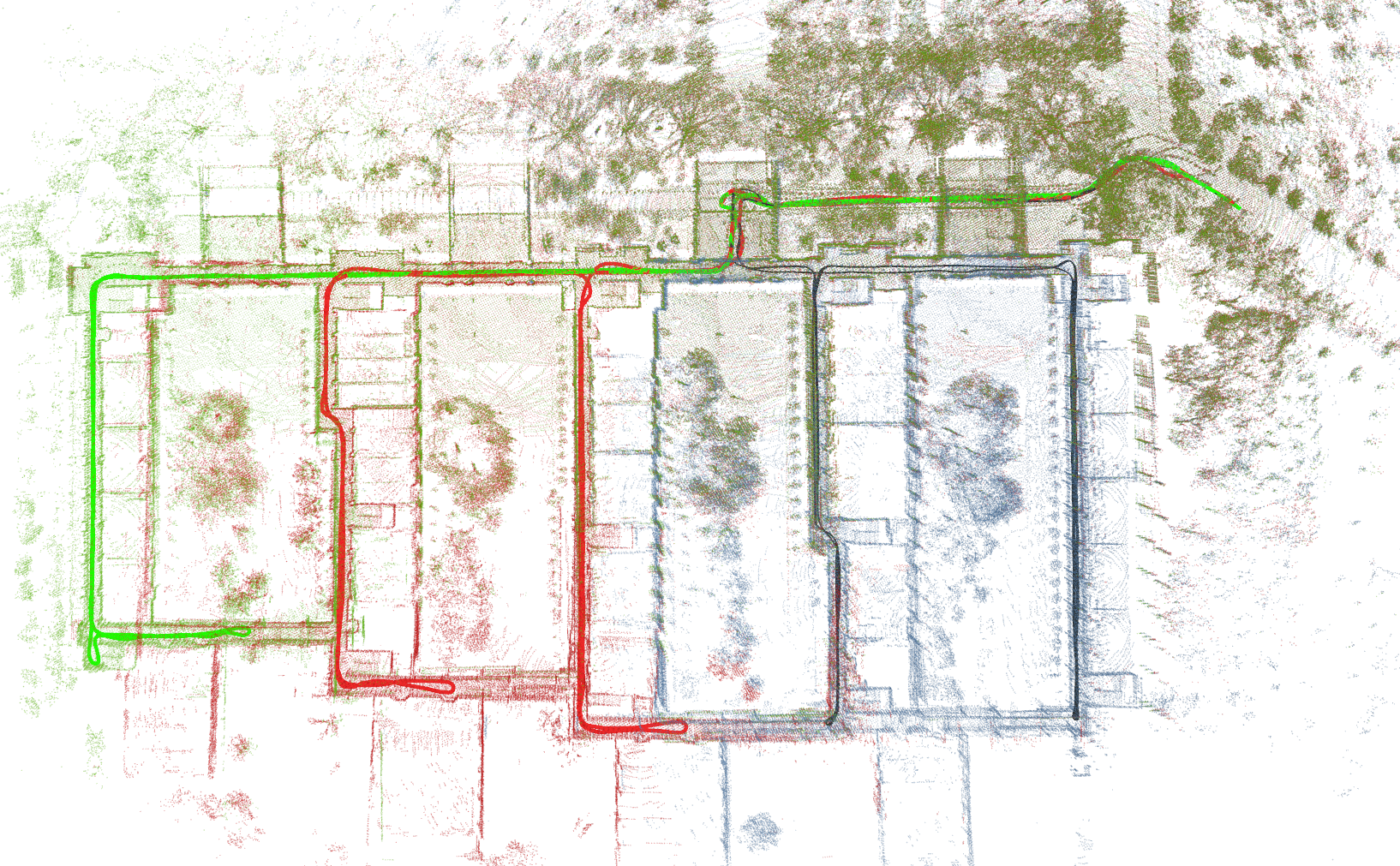}}
	}
	\caption{The trajectories result of DCL-LIO-SAM on \textit{Teaching\_building} sequence, where robot $\alpha$/$\beta$/$\gamma$ is represented by green/red/blue.}
	\label{fig:PCM}
\end{figure}
We also experimented in a teaching building on our campus, an environment with severe perceptual aliasing, with a likelihood threshold of 5.348 and an inlier threshold of 0.3 for geometric verification using RANSAC algorithm.
Fig. \ref{fig:PCM} reports the trajectories and mapping results of DCL-LIO-SAM with and without the PCM approach.
The output trajectories of the robotic swarm with the PCM method in Fig. \ref{fig:withpcm} are more consistent with the surrounding environment than that in Fig. \ref{fig:withoutpcm}, which explains the necessity of introducing an outlier rejection module.

\subsection{Communication}

To quantify the bandwidth requirements of the proposed DCL-SLAM system, we also counted the network traffic on public (HDL-64) and our campus datasets (VLP-16), and the mean bandwidth is shown in Tab. \ref{tab:bandwidth}.
Note that we assume there are no communication range constraints in the robotic swarm, so messages can be exchanged at any time.
For \textit{Library} sequence, the distributed PGO requires the transmission of 3.46 MB, while the centralized one requires 10.06 MB to transmit the whole pose graph.

Furthermore, we counted the sizes of messages sent among swarm on our datasets, as shown in Tab. \ref{tab:dataSize}.
In DCL-SLAM, the transmitted data can be broadly grouped into keyframe, inter-robot match, and distributed PGO.
Firstly, the keyframe message in DCL-SLAM is extracted from the keyframe 3D point cloud.
While a single laser scan from VLP-16 is 1 MB and from HDL-64 is 2MB, the message size needed for the DiSCo-SLAM method for each LiDAR keyframe is around 200kB for VLP-16 while transmitting the keyframe feature cloud.
On the contrary, in DCL-SLAM, the size of the global descriptors of each frame is 29.12kB, with sending filtered raw point cloud for verification (99.92kB/frame for VLP-16) in the loop closure message if necessary.
The inter-robot match messages are mainly the loop measurement and verification laser point cloud.
The last distributed PGO message mainly includes optimization states, the rotation, and pose estimates from the neighbor.
In summary, our proposed DCL-SLAM will significantly reduce the communication burden without sharing raw data.

\begin{table}[tbp]
	\centering
	\caption{Network Traffic of Each Agent Averaged over Sequences}
	\label{tab:bandwidth}
	\begin{center}
		\tabcolsep=0.12cm
		\renewcommand\arraystretch{1.15}
		\begin{tabular}{cccccc}
			\toprule
			\multicolumn{2}{c}{\multirow{2}{*}{$\mathbf{Datasets}$}} & \multicolumn{3}{c}{\multirow{1}{*}{$\mathbf{Avg.\ Transmit\ Data}$ [kB/s]}} & $\mathbf{Communication}$\\
			& & Robot $\alpha$ & Robot $\beta$ & Robot $\gamma$ & $\mathbf{Time}$ [s]\\
			\midrule
			\multirow{4}{*}{\rotatebox{90}{KITTI}} & \textit{Seq. 09} & 92.93&177.30&91.14 & 101.24 \\
			& \textit{Seq. 08} & 161.78&160.93&- & 191.86 \\
			& \textit{Seq. 05} & 81.08&143.41&187.73 & 251.70 \\
			
			\cmidrule{2-6}
			& Avg. (3 Seq.) &  & 164.67 & & 181.60 \\
			\midrule

			\multirow{4}{*}{\rotatebox{90}{Campus}} & \textit{Library} & 34.43&60.51&43.92 & 1014.44 \\
			& \textit{College} & 66.98&67.48&57.98 & 1188.84 \\
			& \textit{Square\_1} & 41.75&40.68&37.89 & 538.99\\
			
			\cmidrule{2-6}
			& Avg. (3 Seq.) & & 53.55 & & 914.09 \\
			\midrule 
		\end{tabular}
	\end{center}
\end{table}
\begin{table}[tbp]
	\centering
	\caption{Data Size of Messages Sent (VLP-16)}
	\label{tab:dataSize}
	\begin{center}
		\tabcolsep=0.12cm
		\renewcommand\arraystretch{1.15}
		\begin{tabular}{ccc}
			\toprule
			\multicolumn{2}{c}{\multirow{1}{*}{$\mathbf{Information\ of\ Message\ Sent}$}}  & $\mathbf{Avg. Size}$ [kB] $\pm$ Std.\\
			\midrule
			\multirow{2}{*}{Keyframe} & Global Descriptor & 29.12 $\pm$ 0.00 \\
			& Feature Cloud* & 160.24 $\pm$ 16.00\\
			\midrule
			\multirow{3}{*}{Inter-robot Match} & Loop Closure Measurement & 0.11 $\pm$ 0.00\\
			& Verification Point Cloud & 99.92 $\pm$ 13.96\\
			& Raw Point Cloud* & 384.76 $\pm$ 30.76\\
			\midrule
			Distributed PGO & Pose Estimate & 0.10 $\pm$ 0.00 \\
			\bottomrule
		\end{tabular}
	\end{center}
	\footnotesize{* The messages sent in case the robots were to directly transmit raw data.}
\end{table}

\subsection{Large-Scare C-SLAM with 9 Robots}
\begin{figure}[t]
	\centering
	\subfloat[Trajectories]
	{
		\label{fig:large-scale traj} 
		{\includegraphics[width=0.53\linewidth]{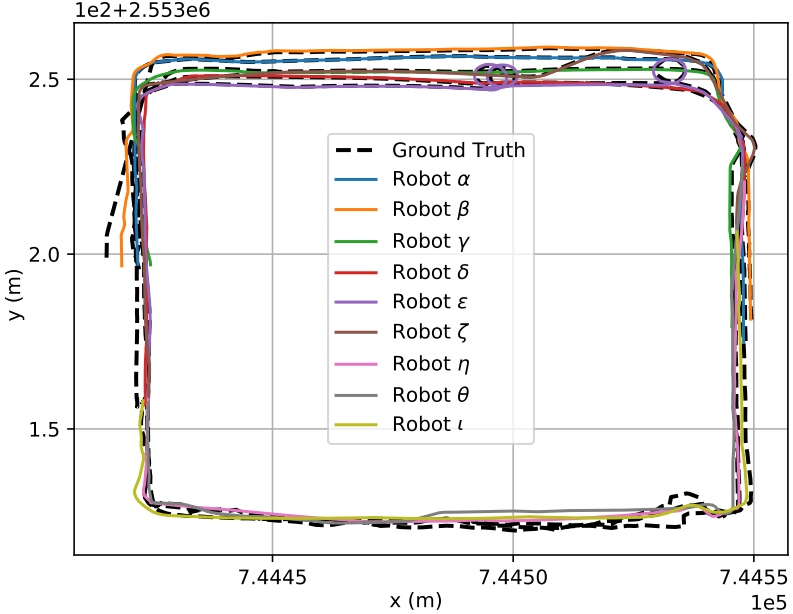}}
	}
	\subfloat[Maps]
	{
		\label{fig:large-scale map} 
		{\includegraphics[width=0.4\linewidth]{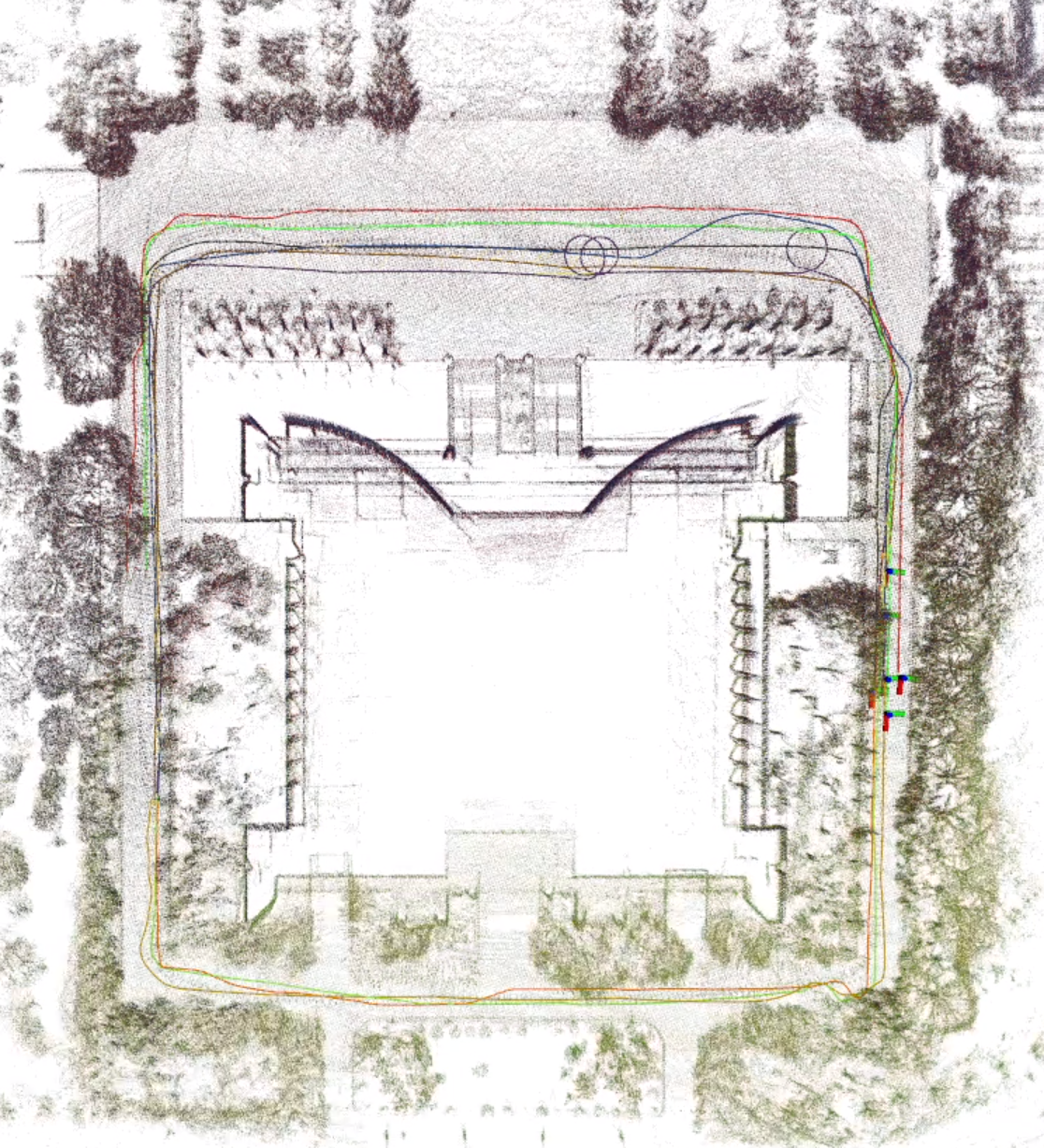}}
	}
	\caption{The trajectories and maps result of DCL-LIO-SAM on a large-scale experiment, where the map of robot $\alpha$/$\beta$/$\gamma$/$\delta$/$\epsilon$/$\zeta$/$\eta$/$\theta$/$\iota$ is represented by green/red/black/brown/purple/blue/dark green/orange/dark yellow. }
	\label{fig:large-scale}
\end{figure}

In this experiment, we evaluate the applicability of DCL-SLAM to a scenario with a large team of UGVs.
Similar to the operation on public datasets, we split sequences collected on Sun Yat-Sen University's eastern campus library to generate a new dataset with 9 UGVs, which provides RTK as ground truth.
As shown in Fig. \ref{fig:large-scale traj}, it comprises 9 U-shaped trajectories of over 250m, covering the whole library building, and Fig. \ref{fig:large-scale map} shows the collaborative mapping generated by DCL-LIO-SAM.
Note that the outlier rejection module is not enabled in this experiment, and it is assumed that the network is full-connected.
The quantization results are as follows: the average ATE of the nine trajectories is 1.83m;
the average bandwidth of transmitted data is 34.36kB/s; the average bandwidth of received data is 193.57 kB/s.
The bandwidth is strongly positively correlated with the number of UGVs because of the end-to-end information sharing.
